\pdfoutput=1
\documentclass[11pt]{article}

\usepackage[]{acl}

\usepackage{times}
\usepackage{latexsym}

\usepackage[T1]{fontenc}
\usepackage[utf8]{inputenc}

\usepackage{microtype}

\usepackage{inconsolata}

\usepackage{amsmath}

\usepackage{multirow}
\usepackage{multicol}
\usepackage{makecell}
\usepackage{graphicx} 
\usepackage{amssymb}
\usepackage{algorithm}
\usepackage{algpseudocode}
\usepackage{ulem}
\usepackage{booktabs}
\usepackage{CJKutf8}
\usepackage{subfigure}

\usepackage{arydshln}
\usepackage{enumitem}
\usepackage{comment}
\usepackage{cleveref}

\usepackage{tikz}
\newcommand*{\circled}[1]{\lower.7ex\hbox{\tikz\draw (0pt, 0pt)%
    circle (.5em) node {\makebox[1em][c]{\small #1}};}}
\robustify{\circled}

\newcommand{\newcheckmark}{\raisebox{0.6ex}{\scalebox{0.7}{$\sqrt{}$}}}
\newcommand{\newcrossmark}{\scalebox{0.85}[1]{$\times$}}

\title{Towards Boosting Many-to-Many Multilingual Machine Translation with Large Language Models}

\author{Pengzhi Gao, Zhongjun He, Hua Wu, and Haifeng Wang \\
Baidu Inc. No. 10, Shangdi 10th Street, Beijing, 100085, China \\
\texttt{\{gaopengzhi,hezhongjun,wu\_hua,wanghaifeng\}@baidu.com} 
}

\begin{document}

\maketitle

\begin{abstract}

The training paradigm for machine translation has gradually shifted, from learning neural machine translation (NMT) models with extensive parallel corpora to instruction finetuning on multilingual large language models (LLMs) with high-quality translation pairs. In this paper, we focus on boosting many-to-many multilingual translation of LLMs with an emphasis on zero-shot translation directions.\footnote{In this work, supervised or zero-shot translation refers to translating between language pairs seen or unseen during multilingual finetuning with translation instructions.} We demonstrate that prompt strategies adopted during finetuning are crucial to zero-shot translation and introduce a cross-lingual consistency regularization, XConST, to bridge the representation gap among different languages and improve zero-shot translation performance. XConST is not a new method, but a version of CrossConST \cite{gao-etal-2023-improving} adapted for translation instruction finetuning with LLMs. Experimental results on ALMA \cite{xu2023paradigm}, Tower \cite{unbabel2024tower}, and LLaMA-2 \cite{touvron2023llama} show that our approach consistently improves translation performance. Our implementations are available at \url{https://github.com/gpengzhi/CrossConST-LLM}.

\end{abstract}

\section{Introduction}

Large language models (LLMs) such as GPT models \cite{openai2023chatgpt,openai2023gpt4}, PaLM models \cite{chowdhery2022palm,anil2023palm}, and others have demonstrated remarkable capabilities in multilingual translation \cite{jiao2023chatgpt,vilar-etal-2023-prompting}. For example, GPT-4 exhibits superior performance over the state-of-the-art (SOTA) multilingual neural machine translation (NMT) models, such as NLLB models \cite{nllbteam2022language}, for high-resource languages \cite{hendy2023good,zhu2023multilingual}. To further boost LLMs' translation capability, the techniques explored by existing approaches include in-context learning \cite{lin-etal-2022-shot,zhu2023multilingual}, continual pretraining with parallel data \cite{yang2023bigtranslate,unbabel2024tower}, and multilingual finetuning with translation instructions \cite{zhang2023bayling,zeng2023tim,li2023eliciting,xu2023paradigm}.

Specifically, \citet{xu2023paradigm} introduce an LLM-based training strategy for multilingual translation, which consists of two stages: continual pretraining on monolingual datasets followed by instruction finetuning on a small set of high-quality translation pairs. By leveraging LLaMA-2 as the foundational model, they learn the ALMA models that outperform NLLB-54.5B and GPT-3.5 across ten translation directions on average. Despite their impressive supervised performance in the English-centric directions, the zero-shot translation evaluation among non-English languages is neglected.

In this paper, our primary goal is to provide a simple yet effective strategy for learning many-to-many multilingual translation with LLMs. Inspired by \citet{wu-etal-2021-language} and \citet{gao-etal-2023-improving}, we investigate the impact of prompt strategies on multilingual translation and show that the zero-shot translation performance varies significantly across different strategies. We then propose a cross-lingual consistency regularization, XConST, for boosting zero-shot translation, where we introduce the explicit constraints to the semantic-equivalent sentence pairs by leveraging Kullback-Leibler (KL) regularization. The contributions of this paper can be summarized as follows:
\begin{itemize}
\item We show that prompt strategies are crucial for zero-shot translation, and there isn't a universal strategy that excels in every scenario.
\item We propose XConST, a simple but effective training strategy that could consistently boost the zero-shot translation performance of LLMs across different prompt strategies.
\item Experimental results on ALMA, Tower, and LLaMA-2 models show that XConST could achieve significant performance improvement on many-to-many multilingual translation.  % on two multilingual machine translation benchmarks.
\end{itemize}

\section{Background}

\subsection{Language Tag for Multilingual NMT}

The multilingual NMT model has a promising capability of zero-shot translation, where it could directly translate between language pairs unseen during training. Language tag is often adopted in multilingual NMT for indicating the translation directions. \citet{wu-etal-2021-language} investigate four widely used language tag strategies, T-ENC, T-DEC, S-ENC-T-ENC, and S-ENC-T-DEC, demonstrated in Table \ref{tab:lang_tag}. They show that language tag is not only the translation direction indicator but also crucial for the zero-shot translation performance. They conclude that ignoring the source language tag and placing the target language tag on the encoder side, which is T-ENC, could enhance the semantics representation consistency among languages and boost the zero-shot translation performance.

\begin{table}[h]\small
\centering
\begin{tabular}{l | l} 
Prompt Strategy & Template \\
\hline
\hline
T-ENC & [tgt]: <SRC>\textbackslash n<TGT>\\
\hline
T-DEC & <SRC>\textbackslash n[tgt]:<TGT> \\
\hline
S-ENC-T-ENC & [src] [tgt]: <SRC>\textbackslash n<TGT> \\
\hline
S-ENC-T-DEC & [src]: <SRC>\textbackslash n[tgt]:<TGT> \\
\hline
\multirow{2}{5em}{GPT-MT} & Translate this from [src] into [tgt]:\textbackslash n \\
 & [src]: <SRC>\textbackslash n[tgt]:<TGT> \\
\end{tabular}
\caption{Different prompt strategies adopted in this paper, where [src] and [tgt] denote the source and target languages, and <SRC> and <TGT> denote the source and target sentences. Note that our examples are slightly different from those in \citet{wu-etal-2021-language} since our strategies are designed for decoder-only LLM while theirs are for NMT model with encoder-decoder architecture.}
\label{tab:lang_tag}
\end{table}

\subsection{Cross-lingual Consistency Regularization for Multilingual NMT}

\citet{gao-etal-2023-improving} introduce a cross-lingual consistency regularization, CrossConST, to bridge the representation gap among different languages and improve zero-shot translation in multilingual NMT. For each sentence pair $(\mathbf{x}, \mathbf{y})$, the training objective of CrossConST is defined as:
\begin{equation}
\mathcal{L}_{CrossConST}(\theta) = \mathcal{L}^{mt}_{ce}(\theta) + \alpha \mathcal{L}^{mt}_{kl}(\theta),
\end{equation}
where
\begin{equation}
\mathcal{L}^{mt}_{ce}(\theta) =  \ell(f(\mathbf{x}, \mathbf{y}; \theta), \ddot{\mathbf{y}}),
\end{equation}
$\ell$ denotes the cross-entropy loss, $\theta$ is a set of model parameters, $f(\mathbf{x}, \mathbf{y}; \theta)$ is a sequence of probability predictions, %i.e., 
%\begin{equation}
%f_j(\mathbf{x}, \mathbf{y}; \theta) = P(y|\mathbf{x}, \mathbf{y}_{<j}; \theta),
%\end{equation}
$\ddot{\mathbf{y}}$ is a sequence of one-hot labels,
\begin{equation}
\mathcal{L}^{mt}_{kl}(\theta) = \text{KL}(f(\mathbf{x}, \mathbf{y}; \theta) \| f(\mathbf{y}, \mathbf{y}; \theta)),
\end{equation}
$\text{KL}(\cdot \| \cdot)$ denotes the Kullback-Leibler (KL) divergence between two distributions, and $\alpha$ is a scalar hyper-parameter that balances $\mathcal{L}^{mt}_{ce}(\theta)$ and $\mathcal{L}^{mt}_{kl}(\theta)$.

\section{Datasets and Baseline Settings}\label{sec:data_model}

In this section, we describe the datasets used in Section \ref{sec:main} as well as the model configurations. % For fair comparisons, we keep our experimental settings consistent with previous work \cite{xu2023paradigm}.

\paragraph{Datasets}

We consider the English-centric scenario and conduct our experiments on five languages: Czech (\texttt{cs}), German (\texttt{de}), English (\texttt{en}), Russian (\texttt{ru}), and Chinese (\texttt{zh}). We collect test data from WMT17 to WMT20 \cite{bojar-etal-2017-findings,bojar-etal-2018-findings,barrault-etal-2019-findings,barrault-etal-2020-findings} and dev data from FLORES-200 \cite{nllbteam2022language} as the training dataset. The detailed statistics of all datasets are summarized in Table \ref{tab:dataset}. For supervised direction evaluation, we collect the test data from WMT22 \cite{kocmi-etal-2022-findings} resulting in $8$ translation directions. For zero-shot direction evaluation, we collect the devtest data from FLORES-200 resulting in $12$ translation directions. We evaluate the model performance by case-sensitive SacreBLEU\footnote{We apply \texttt{zh} tokenizer for Chinese and \texttt{13a} for others.} \cite{post-2018-call} and COMET\footnote{https://huggingface.co/Unbabel/wmt22-comet-da} \cite{rei-etal-2020-comet}.

\paragraph{Settings}

\begin{table*}[h]\small
\centering
\begin{tabular}{l | c | c c c | c c c} 
\multicolumn{1}{c|}{Prompt Strategy} & LoRA & Supervised & Zero-Shot & Pivot & Supervised & Zero-Shot & Pivot \\
 & & \multicolumn{3}{c|}{\textit{Performance on ALMA-7B-Pretrain}} & \multicolumn{3}{c}{\textit{Performance on ALMA-13B-Pretrain}} \\
\hline
\hline
T-ENC & \newcrossmark & 31.6 / 84.8 & 21.2 / 84.5 & 20.6 / {85.4} & 33.0 / 85.4 & {\bf 23.4} / {\bf 86.3} & 21.5 / 85.8 \\
T-DEC & \newcrossmark & {31.7} / 84.9 & {\bf 22.4} / {\bf 86.5} & 20.3 / 85.2 & 32.9 / {85.5} & 19.7 / 81.3 & 21.6 / 85.9 \\
S-ENC-T-ENC & \newcrossmark & 31.6 / 84.8 & 2.4 / 58.7 & {20.9} / {85.4} & {33.1} / {85.5} & 10.6 / 68.0 & 21.5 / 85.9 \\
S-ENC-T-DEC & \newcrossmark & {31.7} / {85.0} & {\bf 22.4} / {\bf 86.0} & 20.7 / {85.4} & 32.8 / 85.4 & {\bf 20.3} / {\bf 84.0} & 21.4 / 85.8 \\
GPT-MT & \newcrossmark & {31.7} / {85.0} & 22.1 / 85.5 & 20.5 / 85.3 & 33.0 / 85.4 & 15.2 / 73.9 & {21.9} / {86.0} \\
\hline
T-ENC & \newcheckmark & 32.1 / 84.6 & 21.9 / 85.4 & 22.6 / 86.1 & 34.4 / 85.6 & {\bf 22.4} / {\bf 85.1} & 24.7 / 86.8 \\
T-DEC & \newcheckmark & {32.5} / {85.2} & {\bf 22.4} / {\bf 86.1} & 22.5 / 86.0 & 34.3 / 85.6 & {\bf 22.6} / {\bf 83.9} & 24.7 / 86.8 \\
S-ENC-T-ENC & \newcheckmark & 32.2 / 84.8 & 19.4 / 83.9 & 22.7 / 86.1 & {34.5} / 85.6 & 10.7 / 66.6 & 24.8 / 86.8 \\
S-ENC-T-DEC & \newcheckmark & {32.5} / 85.1 & 21.7 / 85.2 & {22.8} / {86.2} & {34.5} / {85.7} & 5.1 / 60.5 & {24.9} / {86.9} \\
GPT-MT & \newcheckmark & {32.5} / 85.1 & {\bf 22.0} / {\bf 85.5} & 22.7 / 86.1 & 34.4 / 85.6 & 14.0 / 70.4 & 24.8 / 86.8 \\
\end{tabular}
\caption{Performance (SacreBLEU / COMET) on WMT22 and FLORES-200 benchmarks with full-weight and LoRA vanilla finetuning on ALMA-7B-Pretrain and ALMA-13B-Pretrain. ``LoRA'' denotes whether LoRA is utilized during the finetuning procedure. ``Pivot'' denotes the zero-shot translation performance where we first translate the source language into English before generating the target language. The two highest scores of each setting in the zero-shot directions are marked in bold. The detailed results are summarized in Tables \ref{tab:en-centric_7B_vanilla} and \ref{tab:en-centric_13B_vanilla}.}
\label{tab:en-centric_vanilla}
\end{table*}

We adopt ALMA-7B-Pretrain (and 13B) from \citet{xu2023paradigm}'s release\footnote{https://huggingface.co/haoranxu} as our pretrained models which are full-weight continual pretrained on LLaMA-2-7B (and 13B) with monolingual datasets in six languages\footnote{Czech, German, English, Icelandic, Russian, and Chinese.}. We explore both full-weight and low-rank adaptation (LoRA) \cite{hu2022lora} finetuning in our experiments. When LoRA is utilized, we apply LoRA to the down-projection layer in each feed-forward layer and set the rank to $16$. Note that only $0.1\%$ parameters (7.7M for the 7B model and 12M for the 13B model) are updated when using LoRA. We finetune the pretrained models with a batch size of $256$ and set the maximum sequence length to $512$. We use the AdamW optimizer with learning rates 2e-5 and 1e-4 for full-weight and LoRA finetuning respectively. We apply beam search decoding with beam size $5$. We train all models with llama-recipes\footnote{https://github.com/facebookresearch/llama-recipes} toolkit for one epoch on 8 NVIDIA Tesla A100 GPUs.

\section{Methodology}\label{sec:main}

In this section, we investigate the impact of prompt strategies on many-to-many multilingual translation with LLMs (Section \ref{sec:lt_strategy}) and propose a cross-lingual consistency regularization method for boosting zero-shot translation (Section \ref{sec:xconst}). We introduce the details of each part below.

\subsection{Multilingual Finetuning with Translation Instructions}\label{sec:lt_strategy}

Following the similar lines of \citet{li2023eliciting} and \citet{xu2023paradigm}, we finetune the pretrained LLMs with high-quality parallel datasets as the translation instructions. Consider a decoder-only Transformer \cite{NIPS2017_3f5ee243} model parameterized by $\theta$. Let $\mathbf{x}$ and $\mathbf{y}$ denote the source and target sentences respectively. A fixed prompt template, denoted as $\mathcal{I}$, is applied to guide the model for translating sentences. The training objective is defined as:
\begin{equation}\label{eq:llm_ce}
\mathcal{L}^{llm}_{ce}(\theta) =  \ell(f(\mathbf{x}, \mathbf{y}, \mathcal{I}; \theta), \ddot{\mathbf{y}}),
\end{equation}
where $f(\mathbf{x}, \mathbf{y}, \mathcal{I}; \theta)$ is a sequence of probability predictions, i.e., 
\begin{equation}
f_j(\mathbf{x}, \mathbf{y}, \mathcal{I}; \theta) = P(y|\mathbf{x}, \mathbf{y}_{<j}, \mathcal{I}; \theta).
\end{equation}
Note that \eqref{eq:llm_ce} is a standard causal language modeling (CLM) loss, which predicts the next token in a sequence of tokens, and the model can only attend to tokens on the left. We only compute loss for the target sentence during instruction finetuning.

\subsubsection{Does Prompt Strategy Matter for Multilingual Translation?}\label{sec:exp_vanilla}

\begin{figure*}[h]
\centering
\includegraphics[scale=0.25]{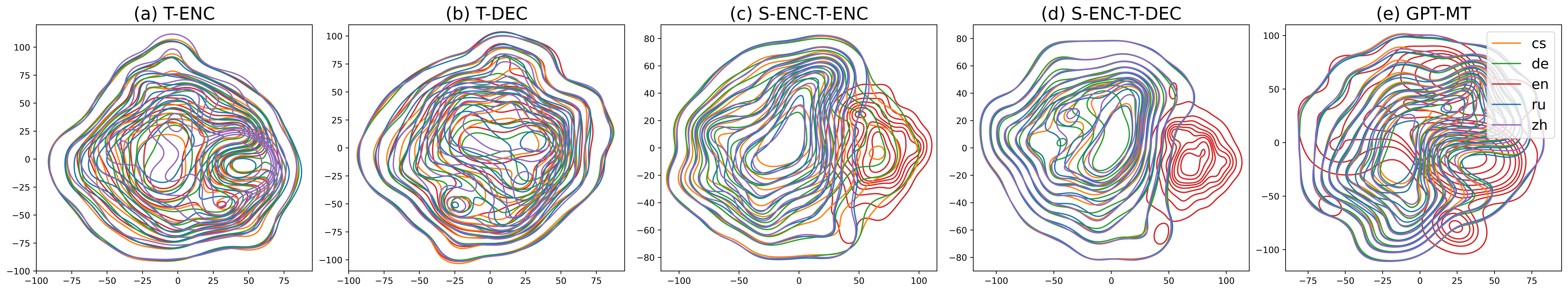}
\caption{Bivariate kernel density estimation plots of sentence representations after using T-SNE dimensionality reduction on the multi-way parallel devtest data from FLORES-200. The orange line denotes Czech, the green line denotes German, the red line denotes English, the blue line denotes Russian, and the purple line denotes Chinese.}
\label{fig:kde_vanilla}
\end{figure*}

We adopt five prompt strategies listed in Table \ref{tab:lang_tag}, where four strategies (T-ENC, T-DEC, S-ENC-T-ENC, and S-ENC-T-DEC) are widely used for learning multilingual NMT, and GPT-MT \cite{hendy2023good} is a popular translation prompt adopted in the LLM field. We report averaged BLEU and COMET scores of all prompt strategies in Table \ref{tab:en-centric_vanilla}. By checking model performance under different combinations of prompt and finetuning strategies, we have the following observations:
\begin{itemize}[leftmargin=*]
\item All prompt strategies achieve comparable performance in the supervised directions. LoRA finetuning slightly outperforms full-weight finetuning, especially regarding the BLEU scores.
\item The zero-shot translation performance varies significantly across different prompt strategies, and there is no single strategy that performs the best for all cases. Although 13B models achieve superior supervised performance, they generally underperform 7B models in the zero-shot directions, which can not be resolved by utilizing LoRA.
\item When prompt strategies are selected properly, the zero-shot translation could achieve comparable or better performance than the pivot-based approach, which is quite reasonable: pretrained LLMs already have translation capability, and translation instructions are only used to elicit it.
\end{itemize}

\subsubsection{Visualization Analysis}\label{sec:visualization}

We here conduct experiments on the multi-way parallel devtest data from FLORES-200, where $1012$ sentences have translations in different languages. We construct the prompts with \texttt{xx} $\rightarrow$ \texttt{ru} translation instruction and calculate the sentence representations by extracting the last Transformer layer outputs of the last token in the input sequence. We utilize the models that are LoRA finetuned on ALMA-13B-Pretrain in Table \ref{tab:en-centric_vanilla} and apply dimension reduction on the 5120-dimensional sentence representations with T-SNE \cite{NIPS2002_6150ccc6} and then depict the bivariate kernel density estimation based on the 2-dimensional representations in Figure \ref{fig:kde_vanilla}. We can see that our models cannot align these five languages well in the representation space. It is worth mentioning that T-ENC and T-DEC align the languages much better than the other three strategies, which is also reflected in the zero-shot translation performance in Table \ref{tab:en-centric_vanilla}.

\subsubsection{Can Off-Target Issue be Mitigated by Instruction Generalization?}

We here apply fastText\footnote{https://fasttext.cc/docs/en/language-identification.html} \cite{joulin2016fasttextzip,joulin-etal-2017-bag} toolkit for language identification and calculate the off-target ratio (Table \ref{tab:en-centric_vanilla_off-target}) in the zero-shot directions for each model demonstrated in Table \ref{tab:en-centric_vanilla}. The off-target ratio is the average percentage of sentences translated into the wrong languages. We can see that the bad translation performance in the zero-shot directions is highly related to the off-target issue which could also be regarded as the failure of LLMs to follow the translation instructions. We then investigate two strategies for improving the instruction-following capability of LLMs.

\begin{table}[h]\small
\centering
\begin{tabular}{l | c | c | c} 
\multicolumn{1}{c|}{Prompt Strategy} & LoRA & Off-Target & Off-Target \\
 & & 7B (\%) & 13B (\%) \\
\hline
\hline
T-ENC & \newcrossmark & 5.94 & 2.33 \\
T-DEC & \newcrossmark & 0.11 & 16.58 \\
S-ENC-T-ENC & \newcrossmark & 92.32 & 60.80 \\
S-ENC-T-DEC & \newcrossmark & 1.38 & 12.78 \\
GPT-MT & \newcrossmark & 2.68 & 38.07 \\
\hline
T-ENC & \newcheckmark & 3.38 & 8.23 \\
T-DEC & \newcheckmark & 1.08 & 10.91 \\
S-ENC-T-ENC & \newcheckmark & 11.08 & 66.11 \\
S-ENC-T-DEC & \newcheckmark & 3.94 & 87.22 \\
GPT-MT & \newcheckmark & 3.10 & 50.96 \\
\end{tabular}
\caption{The off-target ratio in the zero-shot translation directions for each model demonstrated in Table \ref{tab:en-centric_vanilla}.}
\label{tab:en-centric_vanilla_off-target}
\end{table}

\begin{table}[h]\small
\centering
\begin{tabular}{l | c | c} 
\multicolumn{1}{c|}{Method} & Supervised & Zero-shot \\
\hline
\hline
7B w/ full-weight finetune & 31.7 / 85.0 & 22.1 / 85.5 \\
\ \ \ + Diversification & 31.6 / 85.0 & 22.2 / 86.1 \\
\ \ \ + Alpaca dataset & 31.0 / 84.7 & 21.7 / 85.2  \\
\hline
7B w/ LoRA finetune & 32.5 / 85.1 & 22.0 / 85.5 \\
\ \ \ + Diversification & 32.7 / 85.2 & 22.3 / 86.2 \\
\ \ \ + Alpaca dataset & 31.8 / 84.8 & 17.4 / 78.9  \\
\hline
13B w/ full-weight finetune & 33.0 / 85.4 & 15.2 / 73.9 \\
\ \ \ + Diversification & 32.9 / 85.4 & 12.0 / 68.9 \\
\ \ \ + Alpaca dataset & 31.8 / 84.9 & 10.0 / 69.4 \\
\hline
13B w/ LoRA finetune & 34.4 / 85.6 & 14.0 / 70.4 \\
\ \ \ + Diversification & 34.3 / 85.6 & 13.1 / 69.7 \\
\ \ \ + Alpaca dataset & 34.2 / 85.5 & 18.4 / 77.8 \\
\end{tabular}
\caption{Performance (SacreBLEU / COMET) on WMT22 and FLORES-200 benchmarks with different instruction finetuning strategies.}
\label{tab:en-centric_vanilla_extra_results}
\end{table}

\begin{itemize}[leftmargin=*]
\item We diversify the translation instructions, where the prompt strategy is randomly selected from Table \ref{tab:lang_tag} during instruction finetuning.
\item We leverage the additional general instructions, Alpaca dataset \cite{alpaca}, when finetuning on LLMs with translation instructions, where Alpaca is a dataset of 52K instructions generated by OpenAI's text-davinci-003 engine. 
\end{itemize}

We generate translations with GPT-MT prompt strategy and summarize the experimental results in Table \ref{tab:en-centric_vanilla_extra_results}. We can see that instruction diversification could slightly improve the zero-shot translation performance for 7B models but does not work for 13B models. The zero-shot translation can not be consistently improved by introducing general instructions such as Alpaca, and it has a negative impact on the supervised translation performance, which might be due to the decreased ratio of the translation instructions in the finetuning dataset.

\subsection{Cross-lingual Consistency Regularization for Translation Instruction Finetuning}\label{sec:xconst}

\begin{table*}[h]\small
\centering
\begin{tabular}{l | c | c c | c c} 
\multicolumn{1}{c|}{Prompt Strategy} & LoRA & Supervised & Zero-Shot & Supervised & Zero-Shot \\
 & & \multicolumn{2}{c|}{\textit{Performance on ALMA-7B-Pretrain}} & \multicolumn{2}{c}{\textit{Performance on ALMA-13B-Pretrain}} \\
\hline
\hline
T-ENC & \newcrossmark & 31.7 / 84.9 & 22.3 (+1.1) / 86.3 (+1.8) & 33.2 / 85.5 & 23.7 (+0.3) / 86.8 (+0.5) \\
T-DEC & \newcrossmark & 31.6 / 84.9 & 22.3 (-0.1) / 86.5 (+0.0) & 32.8 / 85.4 & 23.7 (+4.0) / 87.0 (+5.7) \\
S-ENC-T-ENC & \newcrossmark & 31.4 / 84.8 & 21.6 (+19.2) / 85.8 (+27.1) & 33.1 / 85.5 & 23.5 (+12.9) / 86.7 (+18.7) \\
S-ENC-T-DEC & \newcrossmark & 31.6 / 85.0 & 22.3 (-0.1) / 86.1 (+0.1) & 32.6 / 85.4 & 24.0 (+3.7) / 87.1 (+3.1) \\
GPT-MT & \newcrossmark & 31.6 / 85.0 & 22.4 (+0.3) / 86.4 (+0.9) & 32.8 / 85.4 & 23.7 (+8.5) / 87.0 (+13.1) \\
\hline
T-ENC & \newcheckmark & 32.1 / 84.7 & 22.0 (+0.1) / 86.1 (+0.7) & 34.3 / 85.7 & 23.7 (+1.3) / 86.5 (+1.4) \\
T-DEC & \newcheckmark & 32.5 / 85.2 & 22.1 (-0.3) / 86.3 (+0.2) & 34.3 / 85.8 & 24.3 (+1.7) / 87.0 (+3.1) \\
S-ENC-T-ENC & \newcheckmark & 31.8 / 84.8 & 21.5 (+2.1) / 85.8 (+1.9) & 34.3 / 85.7 & 24.0 (+13.3) / 86.5 (+19.9) \\
S-ENC-T-DEC & \newcheckmark & 32.5 / 85.1 & 22.4 (+0.7) / 86.3 (+1.1) & 34.1 / 85.7 & 24.0 (+18.9) / 86.7 (+26.2) \\
GPT-MT & \newcheckmark & 32.5 / 85.1 & 22.4 (+0.4) / 86.3 (+0.8) & 34.2 / 85.7 & 23.9 (+9.9) / 86.7 (+16.3) \\
\end{tabular}
\caption{Performance (SacreBLEU / COMET) on WMT22 and FLORES-200 benchmarks with full-weight and LoRA cross-lingual consistency finetuning on ALMA-7B-Pretrain and ALMA-13B-Pretrain. The zero-shot performance changes are pointed out in the parentheses. The detailed results are summarized in Tables \ref{tab:en-centric_7B_xconst} and \ref{tab:en-centric_13B_xconst}.}
\label{tab:en-centric_xconst}
\end{table*}

\begin{figure*}[h]
\centering
\includegraphics[scale=0.25]{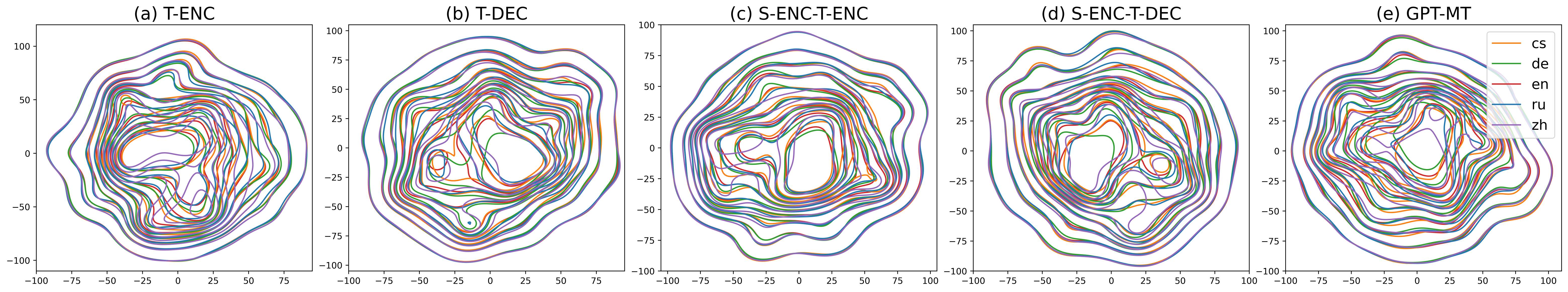}
\caption{Bivariate kernel density estimation plots of sentence representations after using T-SNE dimensionality reduction on the multi-way parallel devtest data from FLORES-200.}
\label{fig:kde_xconst}
\end{figure*}

\begin{figure}[h]
\centering
\includegraphics[scale=0.43]{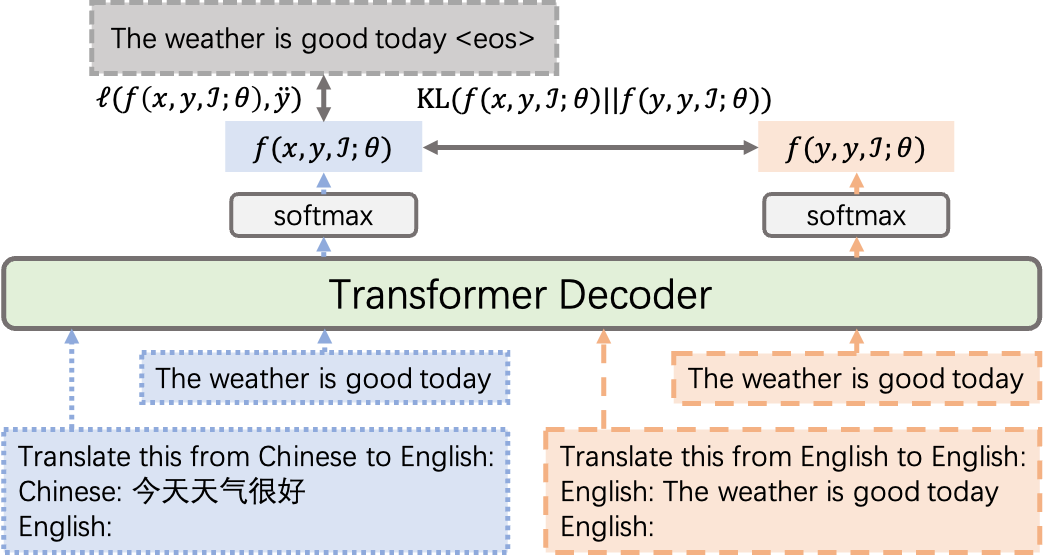}
\caption{Illustration of XConST for translation instruction finetuning, where the original Chinese-English sentence pair ("\begin{CJK}{UTF8}{gbsn}今天天气很好\end{CJK}", "The weather is good today") and the copied English-English sentence pair ("The weather is good today", "The weather is good today") are fed into the multilingual LLM to generate two output distributions $f(\mathbf{x}, \mathbf{y}, \mathcal{I}; \theta)$ and $f(\mathbf{y}, \mathbf{y}, \mathcal{I}; \theta)$.}
\label{fig:crossconst}
\end{figure}

Inspired by CrossConST for multilingual encoder-decoder models \cite{gao-etal-2023-improving}, a natural question arises: \textit{Can we boost decoder-only LLMs' zero-shot translation by leveraging cross-lingual consistency regularization?} The main idea is to close the representation gap among semantic-equivalent sentences in different languages and force the output distribution of the model to be consistent among different semantic-equivalent representations. During finetuning, for each sentence pair $(\mathbf{x}, \mathbf{y})$, the training objective of XConST is defined as:
\begin{equation}\label{main_loss}
\mathcal{L}_{XConST}(\theta) = \mathcal{L}^{llm}_{ce}(\theta) + \alpha \mathcal{L}^{llm}_{kl}(\theta),
\end{equation}
where
\begin{equation}\label{kl_constraint}
\mathcal{L}^{llm}_{kl}(\theta) = \text{KL}(f(\mathbf{x}, \mathbf{y},\mathcal{I}; \theta) \| f(\mathbf{y}, \mathbf{y}, \mathcal{I}; \theta)).
\end{equation}
The gradients are backpropagated through both sides of the KL regularization in XConST. Figure \ref{fig:crossconst} illustrates XConST regularization for finetuning multilingual LLM with translation instructions.

\subsubsection{Experimental Results}\label{sec:exp_crossconst}

Following the experimental setup in Section \ref{sec:exp_vanilla}, we utilize the cross-lingual consistency regularization during translation instruction finetuning and choose $\alpha$ in \eqref{main_loss} from \{0.05, 0.1, 0.25\}. We report averaged BLEU and COMET scores of all prompt strategies in Table \ref{tab:en-centric_xconst}. By checking model performance under different combinations of prompt and finetuning strategies in Tables \ref{tab:en-centric_vanilla} and \ref{tab:en-centric_xconst}, we can see that XConST could achieve strong performance in both supervised and zero-shot translation directions across different prompt strategies. For example, XConST improves the zero-shot translation performance by $3.46$ and $10.8$ COMET scores on average for 7B and 13B models respectively.

Following the experimental setup in Section \ref{sec:visualization}, we visualize the sentence representations of the multi-way parallel devtest data from FLORES-200 in Figure \ref{fig:kde_xconst}. By comparing the representation spaces illustrated in Figures \ref{fig:kde_vanilla} and \ref{fig:kde_xconst}, we can see that the languages cannot be aligned well in the representation space by leveraging vanilla finetuning procedure, while XConST draws the representations across different languages much closer.

\paragraph{Comparison with SOTA Models}

We summarize the translation results of several SOTA models on WMT22 and FLORES-200 benchmarks as follows:
\begin{itemize}[leftmargin=*]
\item {\bf GPT-3.5 Turbo / GPT-4 Turbo}: The translation performance by leveraging the OpenAI API\footnote{https://api.openai.com/v1/chat/completions} with greedy decoding for inference. We use the prompt strategy demonstrated in Table \ref{tab:gpt_prompt}.
\item {\bf NLLB-54.5B}: The largest and best multilingual NMT model with encoder-decoder architecture released by the No Language Left Behind (NLLB) project \cite{nllbteam2022language}.
\item {\bf ALMA-13B-LoRA}: The best translation model released by \citet{xu2023paradigm}, which involves monolingual data finetuning on LLaMA-2 and high-quality parallel data LoRA finetuning.
\end{itemize}
The experimental results are reported in Table \ref{tab:sota}. Our model, which is cross-lingual consistency finetuned on ALMA-13B-Pretrain with LoRA and GPT-MT prompt strategy, substantially outperforms ALMA-13B-LoRA in the zero-shot translation directions. Note that the training data of ALMA-13B-LoRA and ours are almost the same except that they include devtest data from FLORES-200 for instruction finetuning. It is also worth mentioning that our model achieves strong performance on the FLORES-200 benchmark compared with NLLB-54.5B which utilizes extensive parallel sentence pairs in those translation directions.

\begin{table}\small
\centering
\begin{tabular}{l|c|c}
Model & WMT22 & FLORES-200 \\
\hline\hline
GPT-4 Turbo & 36.7 / 87.2 & 27.4 / 88.3 \\
GPT-3.5 Turbo & 34.8 / 86.2 & 24.6 / 86.9 \\
NLLB-54.5B & 31.5 / 81.9 & 24.5 / 86.2 \\
ALMA-13B-LoRA & 34.4 / 85.7 & 14.8 / 69.8 \\
\hline
Ours & 34.2 / 85.7 & 23.9 / 86.7 \\
\end{tabular}
\caption{The translation performance of the SOTA models on WMT22 and FLORES-200 benchmarks.}
\label{tab:sota}
\end{table}

\subsubsection{Effect of $\alpha$}

\begin{figure}[h]
\centering
\includegraphics[scale=0.6]{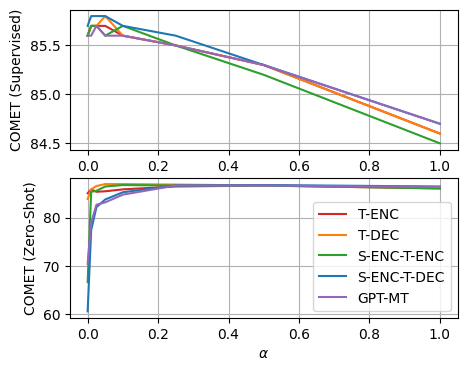}
\caption{The averaged COMET scores with different $\alpha$ on WMT22 and FLORES-200 benchmarks.}
\label{fig:alpha}
\end{figure}

We here investigate the impact of the scalar hyperparameter $\alpha$ in XConST. $\alpha$ is a penalty parameter that controls the regularization strength in our optimization problem. We vary $\alpha$ in \{$0.0$, $0.01$, $0.025$, $0.05$, $0.1$, $0.25$, $0.5$, $1.0$\} and conduct the experiments with LoRA cross-lingual consistency finetuning on ALMA-13B-Pretrain. Note that cross-lingual consistency finetuning is simplified to vanilla finetuning when $\alpha=0.0$. The averaged COMET scores are illustrated in Figure \ref{fig:alpha} due to their better alignment with human evaluations \cite{freitag-etal-2022-results}. We can see that a small $\alpha$ (e.g. $0.1$) is enough to consistently boost the translation performance in the zero-shot directions across different prompt strategies. Meanwhile, an overwhelming regularization ($\alpha=1.0$) is not plausible for the supervised translation performance.

\subsubsection{Does Cross-lingual Consistency Still Work Beyond English-centric Scenario?}

\begin{table}[h]\small
\centering
\begin{tabular}{l | c | c c} 
\multicolumn{1}{c|}{Method} & WMT22 & \multicolumn{2}{c}{FLORES-200} \\
 & & \texttt{cs} $\leftrightarrow$ \texttt{de} & others \\
\hline
\hline
T-ENC & 34.6 / 85.6 & 25.9 / 89.2 & 23.9 / 86.1 \\
\ \ + XConST & 34.5 / 85.7 & 25.8 / 89.2 & 24.2 / 86.6 \\
\hline
T-DEC & 34.5 / 85.7 & 25.9 / 89.3 & 16.9 / 73.4 \\
\ \ + XConST & 34.4 / 85.7 & 25.6 / 89.2 & 23.6 / 86.4 \\
\hline
S-ENC-T-ENC & 34.4 / 85.5 & 25.8 / 89.1 & 24.0 / 86.2 \\
\ \ + XConST & 34.4 / 85.7  & 25.9 / 89.2 & 24.3 / 86.7 \\
\hline
S-ENC-T-DEC & 34.5 / 85.6 & 25.6 / 89.1 & 17.3 / 76.0 \\
\ \ + XConST & 34.3 / 85.7 & 25.5 / 89.1 & 23.8 / 86.4 \\
\hline
GPT-MT &  34.3 / 85.7 & 25.8 / 89.1 & 17.8 / 74.6 \\
\ \ + XConST & 34.3 / 85.7 & 25.6 / 89.2 & 23.9 / 86.4 \\
\end{tabular}
\caption{Performance on the WMT22 and FLORES-200 benchmarks with LoRA cross-lingual consistency finetuning on ALMA-13B-Pretrain.}
\label{tab:beyonden}
\end{table}

We here extend our experiments beyond the English-centric scenario. Specifically, we gather the English-centric datasets used in Section \ref{sec:exp_crossconst} and supplement it with an additional $22990$ \texttt{cs} $\leftrightarrow$ \texttt{de} sentence pairs, which are collected from WMT10 to WMT13 \cite{callison-burch-etal-2010-findings,callison-burch-etal-2011-findings,callison-burch-etal-2012-findings,bojar-etal-2013-findings} test datasets. FLORES-200 benchmark then includes two supervised directions and ten zero-shot directions. We report averaged BLEU and COMET scores of the baseline and our approach in Table \ref{tab:beyonden}. By checking model performance under different combinations of datasets and training strategies in Tables \ref{tab:en-centric_vanilla}, \ref{tab:en-centric_xconst} and \ref{tab:beyonden}, we can see that adding beyond the English-centric dataset usually could improve the overall zero-shot translation performance except for T-DEC in our experiments. The cross-lingual consistency regularization is complementary to the data-based method and could further improve the zero-shot translation performance.

\section{Experiments on More Languages}\label{sec:extend}

To ensure the generalization of our findings, we extend our analysis to a broader range of languages and investigate many-to-many machine translation performance with the Tower and LLaMA-2 models. We utilize LoRA finetuning in our experiments and apply the same training and inference configurations discussed in Section \ref{sec:data_model}. We evaluate the model performance on the FLORES-200 devtest datasets.

\subsection{Experiments with Tower Model}

\paragraph{Dataset Description and Model Configuration}

We extract the English-centric translation pairs from TowerBlocks\footnote{https://huggingface.co/datasets/Unbabel/TowerBlocks-v0.1} dataset, which includes data for multiple translation-related tasks, such as automatic post edition, error span prediction, named-entity recognition, machine translation, and its different variants. The detailed statistics of the translation pairs are summarized in Table \ref{tab:dataset_tower}. We conduct our experiments on ten languages\footnote{Chinese (\texttt{zh}), Dutch (\texttt{nl}), English (\texttt{en}), French (\texttt{fr}), German (\texttt{de}), Italian (\texttt{it}), Korean (\texttt{ko}), Portuguese (\texttt{pt}), Russian (\texttt{ru}), and Spanish (\texttt{es})} and adopt TowerBase-7B\footnote{https://huggingface.co/Unbabel/TowerBase-7B-v0.1} as our pretrained model, which is full-weight continual pretrained on LLaMA-2-7B with monolingual and parallel datasets in those ten languages. We fix $\alpha$ in \eqref{main_loss} to be $0.25$.

\paragraph{Main Results}

\begin{table}[h]\small
\centering
\begin{tabular}{l | c | c | c} 
\multicolumn{1}{c|}{Method} & Data & English-centric & Others \\
\hline
\hline
TowerInstruct & \circled{1} & 37.0 / 88.6 & 24.6 / 86.1 \\
\hline
T-ENC & \circled{2} & 37.4 / 88.5 & 23.8 / 85.2 \\
\ \ + XConST & \circled{2} & 37.6 / 88.6 & 25.0 / 86.0 \\
\hdashline
T-DEC & \circled{2} & 37.6 / 88.6 & 24.5 / 85.8 \\
\ \ + XConST & \circled{2} & 37.7 / 88.7 & 25.2 / 86.2 \\
\hdashline
S-ENC-T-ENC & \circled{2} & 37.5 / 88.6 & 21.6 / 82.8 \\
\ \ + XConST & \circled{2} & 37.7 / 88.6 & 25.1 / 86.1 \\
\hdashline
S-ENC-T-DEC & \circled{2} & 37.8 / 88.6 & 24.0 / 85.2 \\
\ \ + XConST & \circled{2} & 37.8 / 88.7 & 25.3 / 86.2 \\
\hdashline
GPT-MT & \circled{2} & 38.0 / 88.7 & 24.4 / 85.7 \\
\ \ + XConST & \circled{2} & 38.0 / 88.7 & 25.2 / 86.2 \\
\end{tabular}
\caption{Performance (SacreBLEU / COMET) on FLORES-200 benchmark with LoRA vanilla and cross-lingual consistency finetuning on TowerBase-7B. \circled{1} denotes the TowerBlocks dataset. \circled{2} denotes the English-centric translation pairs extracted from TowerBlock. The results with all translation pairs in TowerBlock are summarized in Table \ref{tab:tower_results_beyond}.}
\label{tab:tower_results}
\end{table}

We report averaged BLEU and COMET scores of all prompt and finetuning strategies in Table \ref{tab:tower_results}, where we also include the experimental results of TowerInstruct\footnote{https://huggingface.co/Unbabel/TowerInstruct-7B-v0.1}, a language model that results from finetuning TowerBase-7B on the TowerBlock dataset. We can see that XConST consistently improves the zero-shot translation performance across different prompt strategies and performs comparably or even better than TowerInstruct. It is worth mentioning that TowerInstruct utilizes translation pairs beyond English-centric direction and sophisticated instruction diversification mechanisms, while ours adopts a fixed translation prompt template on the English-centric pairs with cross-lingual consistency regularization.

\subsection{Experiments with LLaMA-2 Model}

\paragraph{Dataset Description and Model Configuration}

\begin{table*}[h]\small
\centering
\begin{tabular}{l | c c c | c c c} 
\multicolumn{1}{c|}{Prompt Strategy} & \texttt{xx} $\leftrightarrow$ \texttt{en} & \texttt{xx} $\leftrightarrow$ \texttt{es} & \texttt{xx} $\leftrightarrow$ \texttt{ru} & \texttt{xx} $\leftrightarrow$ \texttt{en} & \texttt{xx} $\leftrightarrow$ \texttt{es} & \texttt{xx} $\leftrightarrow$ \texttt{ru} \\
 & \multicolumn{3}{c|}{\textit{Performance on LLaMA-2-7B}} & \multicolumn{3}{c}{\textit{Performance on LLaMA-2-13B}} \\
\hline
\hline
T-ENC & 29.6 / 82.2 & 15.1 / 76.8 & 14.9 / 78.4 & 32.0 / 83.6 & 14.6 / 76.1 & 13.6 / 75.7 \\
\ \ + XConST & 29.6 / 82.4 & 16.2 / 78.5 & 15.9 / 80.5 & 31.9 / 83.8 & 17.8 / 80.1 & 18.0 / 82.2 \\
\hline
T-DEC & 29.7 / 82.2 & 16.4 / 78.7 & 16.6 / 80.8 & 31.8 / 83.5 & 17.8 / 80.3 & 18.4 / 82.6 \\
\ \ + XConST & 29.6 / 82.4 & 16.7 / 79.2 & 16.5 / 81.3 & 31.8 / 83.6 & 18.3 / 80.8 & 18.5 / 83.0 \\
\hline
S-ENC-T-ENC & 29.7 / 82.3 & 4.2 / 65.5 & 3.3 / 58.4 & 32.1 / 83.8 & 11.3 / 73.9 & 10.9 / 71.1 \\
\ \ + XConST & 29.6 / 82.5 & 15.9 / 78.5 & 15.4 / 79.5 & 32.1 / 83.9 & 17.8 / 80.5 & 18.1 / 82.1 \\
\hline
S-ENC-T-DEC & 29.9 / 82.3 & 15.5 / 78.0 & 14.2 / 75.6 & 32.0 / 83.7 & 13.5 / 76.6 & 14.8 / 74.4 \\
\ \ + XConST & 29.8 / 82.4 & 17.0 / 79.4 & 16.8 / 81.5 & 31.9 / 83.8 & 18.3 / 80.9 & 18.9 / 83.2 \\
\hline
GPT-MT & 29.9 / 82.3 & 15.7 / 78.3 & 13.4 / 74.1 & 32.1 / 83.8 & 17.5 / 80.2 & 16.1 / 77.1 \\
\ \ + XConST & 29.8 / 82.4 & 16.8 / 79.3 & 16.8 / 81.6 & 32.0 / 83.8 & 18.3 / 80.9 & 18.8 / 83.1 \\
\end{tabular}
\caption{Performance (SacreBLEU / COMET) on FLORES-200 benchmark with LoRA vanilla and cross-lingual consistency finetuning on LLaMA-2-7B and LLaMA-2-13B.}
\label{tab:opus}
\end{table*}

We select $28$ languages\footnote{Afrikaans (\texttt{af}), Asturian (\texttt{ast}), Belarusian (\texttt{be}), Bosnian (\texttt{bs}), Bulgarian (\texttt{bg}), Catalan (\texttt{ca}), Croatian (\texttt{hr}), Czech (\texttt{cs}), Danish (\texttt{da}), Dutch (\texttt{nl}), English (\texttt{en}), French (\texttt{fr}), Galician (\texttt{gl}), German (\texttt{de}), Icelandic (\texttt{is}), Luxembourgish (\texttt{lb}), Macedonian (\texttt{mk}), Norwegian (\texttt{nb}), Occitan (\texttt{oc}), Polish (\texttt{pl}), Portuguese (\texttt{pt}), Romanian (\texttt{ro}), Russian (\texttt{ru}), Serbian (\texttt{sr}), Slovak (\texttt{sk}), Slovenian (\texttt{sl}), Spanish (\texttt{es}), and Ukrainian (\texttt{uk}).} from three different language families (Indo-European-Germanic, Indo-European-Romance, and Indo-European-Slavic), where the LLaMA-2 models achieve impressive few-shot performance in the English-centric translation directions \cite{zhu2023multilingual}. We collect 10K English-centric parallel sentences from the OPUS collection \cite{TIEDEMANN12.463} for each language by adopting the data cleaning process as follows: 1) We remove duplicate sentence pairs and also discard sentence pairs wherein the English sentences exceed $2000$ characters. 2) Language identification filtering is applied by utilizing the fastText toolkit. 3) Semantic similarity filtering is performed based on the MuSR model \cite{gao-etal-2023-learning-multilingual}. In summary, we collect 540K cleaned English-centric sentence pairs covering $28$ languages including English. Note that the same sentence pairs are shared by \texttt{xx} $\leftarrow$ \texttt{en} and \texttt{xx} $\rightarrow$ \texttt{en} directions. Due to limited computational resources, we only generate Spanish-centric and Russian-centric translations for zero-shot evaluation. We adopt LLaMA-2-7B (and 13B) as our pretrained models and fix $\alpha$ in \eqref{main_loss} to be $0.1$.

\paragraph{Main Results}

We report averaged BLEU and COMET scores of all prompt and finetuning strategies in Table \ref{tab:opus}. All prompt strategies achieve comparable performance in the supervised directions, while the zero-shot translation performance varies significantly. Specifically, S-ENC-T-ENC usually performs the worst in our experiments. XConST consistently boosts the zero-shot translation performance across different prompt strategies. For example, XConST improves the \texttt{xx} $\leftrightarrow$ \texttt{ru} translation performance by $7.40$ and $6.56$ COMET scores on average for 7B and 13B models respectively.

\section{Related Work}

Recent works on LLMs demonstrate their strong capabilities in multilingual translation \cite{jiao2023chatgpt,hendy2023good,zhu2023multilingual}. % For example, \citet{zhu2023multilingual} thoroughly evaluate the multilingual translation performance of popular LLMs on the FLORES-101 benchmark \cite{goyal-etal-2022-flores} and compare them with SOTA translation engines, such as NLLB and Google Translate. 
To further improve the translation performance, one direction is to find good in-context learning recipes for machine translation with LLMs \cite{lin-etal-2022-shot,agrawal-etal-2023-context,zhang2023prompting,zhu2023multilingual}. For example, \citet{zhang2023prompting} study how to prompt LLMs for machine translation and find that prompt demonstrations and templates have a substantial impact on translation qualities.

Another direction is to utilize parallel datasets for continual pretraining \cite{yang2023bigtranslate,unbabel2024tower} and translation instruction finetuning \cite{zhang2023bayling,li2023eliciting,xu2023paradigm}. % Specifically, \citet{yang2023bigtranslate} continue training LLaMA with massive Chinese monolingual data and multilingual parallel data followed by multilingual translation instruction finetuning. 
Specifically, \citet{li2023eliciting} focus on eliciting LLMs' translation capability via multilingual finetuning with translation instructions. Although the zero-shot translation performance is investigated, they adopt XGLM-7B \cite{lin-etal-2022-shot} as the foundation model which has relatively weak multilingual capability \cite{zhu2023multilingual}, and the translation performance is far from satisfactory. \citet{xu2023paradigm} propose a two-stage training strategy for multilingual translation with LLMs and outperforms the SOTA translation engines, such as NLLB-54.5B and GPT-3.5. They however mainly focus on the supervised English-centric directions, and the translation performance among non-English languages is unclear. Our work follows this line and has an emphasis on the zero-shot translation directions. We investigate the impact of the prompt strategies on the zero-shot translation performance and introduce a simple yet effective cross-lingual regularization constraint, which effectively reduces discrepancies in representations across languages.

\section{Conclusion}

In this paper, we investigate the impact of prompt strategies on many-to-many multilingual machine translation with LLMs and propose a simple but effective cross-lingual consistency regularization method for multilingual finetuning with translation instructions. Through extensive experiments and visualization analysis, we find that: 1) Different prompt strategies achieve comparable performance in the supervised translation directions, while the zero-shot translation performance varies significantly using different prompt strategies. 2) The zero-shot translation performance could be consistently improved by leveraging cross-lingual consistency regularization during translation instruction finetuning across different prompt strategies. For future work, we will explore the effectiveness of the cross-lingual consistency regularization approach on cross-lingual generalization of LLMs across a wide range of tasks and languages.

\section*{Limitations}

In this paper, we mainly focus on evaluating our approach on the English-centric scenario and beyond. Future research could consider more multilingual machine translation benchmarks with different numbers of languages and training samples and conduct experiments on more challenging training scenarios such as chain configurations where we have multiple bridge languages and different zero-shot distances.

\bibliography{anthology,custom}

\begin{thebibliography}{41}
\expandafter\ifx\csname natexlab\endcsname\relax\def\natexlab#1{#1}\fi

\bibitem[{Agrawal et~al.(2023)Agrawal, Zhou, Lewis, Zettlemoyer, and Ghazvininejad}]{agrawal-etal-2023-context}
Sweta Agrawal, Chunting Zhou, Mike Lewis, Luke Zettlemoyer, and Marjan Ghazvininejad. 2023.
\newblock \href {https://doi.org/10.18653/v1/2023.findings-acl.564} {In-context examples selection for machine translation}.
\newblock In \emph{Findings of the Association for Computational Linguistics: ACL 2023}, pages 8857--8873, Toronto, Canada. Association for Computational Linguistics.

\bibitem[{Anil et~al.(2023)Anil, Dai, Firat, Johnson, Lepikhin, Passos, Shakeri, Taropa, Bailey, Chen, Chu, Clark, Shafey, Huang, Meier-Hellstern, Mishra, Moreira, Omernick, Robinson, Ruder, Tay, Xiao, Xu, Zhang, Abrego, Ahn, Austin, Barham, Botha, Bradbury, Brahma, Brooks, Catasta, Cheng, Cherry, Choquette-Choo, Chowdhery, Crepy, Dave, Dehghani, Dev, Devlin, Díaz, Du, Dyer, Feinberg, Feng, Fienber, Freitag, Garcia, Gehrmann, Gonzalez, Gur-Ari, Hand, Hashemi, Hou, Howland, Hu, Hui, Hurwitz, Isard, Ittycheriah, Jagielski, Jia, Kenealy, Krikun, Kudugunta, Lan, Lee, Lee, Li, Li, Li, Li, Li, Lim, Lin, Liu, Liu, Maggioni, Mahendru, Maynez, Misra, Moussalem, Nado, Nham, Ni, Nystrom, Parrish, Pellat, Polacek, Polozov, Pope, Qiao, Reif, Richter, Riley, Ros, Roy, Saeta, Samuel, Shelby, Slone, Smilkov, So, Sohn, Tokumine, Valter, Vasudevan, Vodrahalli, Wang, Wang, Wang, Wang, Wieting, Wu, Xu, Xu, Xue, Yin, Yu, Zhang, Zheng, Zheng, Zhou, Zhou, Petrov, and Wu}]{anil2023palm}
Rohan Anil, Andrew~M. Dai, Orhan Firat, Melvin Johnson, Dmitry Lepikhin, Alexandre Passos, Siamak Shakeri, Emanuel Taropa, Paige Bailey, Zhifeng Chen, Eric Chu, Jonathan~H. Clark, Laurent~El Shafey, Yanping Huang, Kathy Meier-Hellstern, Gaurav Mishra, Erica Moreira, Mark Omernick, Kevin Robinson, Sebastian Ruder, Yi~Tay, Kefan Xiao, Yuanzhong Xu, Yujing Zhang, Gustavo~Hernandez Abrego, Junwhan Ahn, Jacob Austin, Paul Barham, Jan Botha, James Bradbury, Siddhartha Brahma, Kevin Brooks, Michele Catasta, Yong Cheng, Colin Cherry, Christopher~A. Choquette-Choo, Aakanksha Chowdhery, Clément Crepy, Shachi Dave, Mostafa Dehghani, Sunipa Dev, Jacob Devlin, Mark Díaz, Nan Du, Ethan Dyer, Vlad Feinberg, Fangxiaoyu Feng, Vlad Fienber, Markus Freitag, Xavier Garcia, Sebastian Gehrmann, Lucas Gonzalez, Guy Gur-Ari, Steven Hand, Hadi Hashemi, Le~Hou, Joshua Howland, Andrea Hu, Jeffrey Hui, Jeremy Hurwitz, Michael Isard, Abe Ittycheriah, Matthew Jagielski, Wenhao Jia, Kathleen Kenealy, Maxim Krikun, Sneha Kudugunta, Chang
  Lan, Katherine Lee, Benjamin Lee, Eric Li, Music Li, Wei Li, YaGuang Li, Jian Li, Hyeontaek Lim, Hanzhao Lin, Zhongtao Liu, Frederick Liu, Marcello Maggioni, Aroma Mahendru, Joshua Maynez, Vedant Misra, Maysam Moussalem, Zachary Nado, John Nham, Eric Ni, Andrew Nystrom, Alicia Parrish, Marie Pellat, Martin Polacek, Alex Polozov, Reiner Pope, Siyuan Qiao, Emily Reif, Bryan Richter, Parker Riley, Alex~Castro Ros, Aurko Roy, Brennan Saeta, Rajkumar Samuel, Renee Shelby, Ambrose Slone, Daniel Smilkov, David~R. So, Daniel Sohn, Simon Tokumine, Dasha Valter, Vijay Vasudevan, Kiran Vodrahalli, Xuezhi Wang, Pidong Wang, Zirui Wang, Tao Wang, John Wieting, Yuhuai Wu, Kelvin Xu, Yunhan Xu, Linting Xue, Pengcheng Yin, Jiahui Yu, Qiao Zhang, Steven Zheng, Ce~Zheng, Weikang Zhou, Denny Zhou, Slav Petrov, and Yonghui Wu. 2023.
\newblock \href {http://arxiv.org/abs/2305.10403} {Palm 2 technical report}.

\bibitem[{Barrault et~al.(2020)Barrault, Biesialska, Bojar, Costa-juss{\`a}, Federmann, Graham, Grundkiewicz, Haddow, Huck, Joanis, Kocmi, Koehn, Lo, Ljube{\v{s}}i{\'c}, Monz, Morishita, Nagata, Nakazawa, Pal, Post, and Zampieri}]{barrault-etal-2020-findings}
Lo{\"\i}c Barrault, Magdalena Biesialska, Ond{\v{r}}ej Bojar, Marta~R. Costa-juss{\`a}, Christian Federmann, Yvette Graham, Roman Grundkiewicz, Barry Haddow, Matthias Huck, Eric Joanis, Tom Kocmi, Philipp Koehn, Chi-kiu Lo, Nikola Ljube{\v{s}}i{\'c}, Christof Monz, Makoto Morishita, Masaaki Nagata, Toshiaki Nakazawa, Santanu Pal, Matt Post, and Marcos Zampieri. 2020.
\newblock \href {https://aclanthology.org/2020.wmt-1.1} {Findings of the 2020 conference on machine translation ({WMT}20)}.
\newblock In \emph{Proceedings of the Fifth Conference on Machine Translation}, pages 1--55, Online. Association for Computational Linguistics.

\bibitem[{Barrault et~al.(2019)Barrault, Bojar, Costa-juss{\`a}, Federmann, Fishel, Graham, Haddow, Huck, Koehn, Malmasi, Monz, M{\"u}ller, Pal, Post, and Zampieri}]{barrault-etal-2019-findings}
Lo{\"\i}c Barrault, Ond{\v{r}}ej Bojar, Marta~R. Costa-juss{\`a}, Christian Federmann, Mark Fishel, Yvette Graham, Barry Haddow, Matthias Huck, Philipp Koehn, Shervin Malmasi, Christof Monz, Mathias M{\"u}ller, Santanu Pal, Matt Post, and Marcos Zampieri. 2019.
\newblock \href {https://doi.org/10.18653/v1/W19-5301} {Findings of the 2019 conference on machine translation ({WMT}19)}.
\newblock In \emph{Proceedings of the Fourth Conference on Machine Translation (Volume 2: Shared Task Papers, Day 1)}, pages 1--61, Florence, Italy. Association for Computational Linguistics.

\bibitem[{Bojar et~al.(2013)Bojar, Buck, Callison-Burch, Federmann, Haddow, Koehn, Monz, Post, Soricut, and Specia}]{bojar-etal-2013-findings}
Ond{\v{r}}ej Bojar, Christian Buck, Chris Callison-Burch, Christian Federmann, Barry Haddow, Philipp Koehn, Christof Monz, Matt Post, Radu Soricut, and Lucia Specia. 2013.
\newblock \href {https://aclanthology.org/W13-2201} {Findings of the 2013 {W}orkshop on {S}tatistical {M}achine {T}ranslation}.
\newblock In \emph{Proceedings of the Eighth Workshop on Statistical Machine Translation}, pages 1--44, Sofia, Bulgaria. Association for Computational Linguistics.

\bibitem[{Bojar et~al.(2017)Bojar, Chatterjee, Federmann, Graham, Haddow, Huang, Huck, Koehn, Liu, Logacheva, Monz, Negri, Post, Rubino, Specia, and Turchi}]{bojar-etal-2017-findings}
Ond{\v{r}}ej Bojar, Rajen Chatterjee, Christian Federmann, Yvette Graham, Barry Haddow, Shujian Huang, Matthias Huck, Philipp Koehn, Qun Liu, Varvara Logacheva, Christof Monz, Matteo Negri, Matt Post, Raphael Rubino, Lucia Specia, and Marco Turchi. 2017.
\newblock \href {https://doi.org/10.18653/v1/W17-4717} {Findings of the 2017 conference on machine translation ({WMT}17)}.
\newblock In \emph{Proceedings of the Second Conference on Machine Translation}, pages 169--214, Copenhagen, Denmark. Association for Computational Linguistics.

\bibitem[{Bojar et~al.(2018)Bojar, Federmann, Fishel, Graham, Haddow, Huck, Koehn, and Monz}]{bojar-etal-2018-findings}
Ond{\v{r}}ej Bojar, Christian Federmann, Mark Fishel, Yvette Graham, Barry Haddow, Matthias Huck, Philipp Koehn, and Christof Monz. 2018.
\newblock \href {https://doi.org/10.18653/v1/W18-6401} {Findings of the 2018 conference on machine translation ({WMT}18)}.
\newblock In \emph{Proceedings of the Third Conference on Machine Translation: Shared Task Papers}, pages 272--303, Belgium, Brussels. Association for Computational Linguistics.

\bibitem[{Callison-Burch et~al.(2010)Callison-Burch, Koehn, Monz, Peterson, Przybocki, and Zaidan}]{callison-burch-etal-2010-findings}
Chris Callison-Burch, Philipp Koehn, Christof Monz, Kay Peterson, Mark Przybocki, and Omar Zaidan. 2010.
\newblock \href {https://aclanthology.org/W10-1703} {Findings of the 2010 joint workshop on statistical machine translation and metrics for machine translation}.
\newblock In \emph{Proceedings of the Joint Fifth Workshop on Statistical Machine Translation and {M}etrics{MATR}}, pages 17--53, Uppsala, Sweden. Association for Computational Linguistics.

\bibitem[{Callison-Burch et~al.(2012)Callison-Burch, Koehn, Monz, Post, Soricut, and Specia}]{callison-burch-etal-2012-findings}
Chris Callison-Burch, Philipp Koehn, Christof Monz, Matt Post, Radu Soricut, and Lucia Specia. 2012.
\newblock \href {https://aclanthology.org/W12-3102} {Findings of the 2012 workshop on statistical machine translation}.
\newblock In \emph{Proceedings of the Seventh Workshop on Statistical Machine Translation}, pages 10--51, Montr{\'e}al, Canada. Association for Computational Linguistics.

\bibitem[{Callison-Burch et~al.(2011)Callison-Burch, Koehn, Monz, and Zaidan}]{callison-burch-etal-2011-findings}
Chris Callison-Burch, Philipp Koehn, Christof Monz, and Omar Zaidan. 2011.
\newblock \href {https://aclanthology.org/W11-2103} {Findings of the 2011 workshop on statistical machine translation}.
\newblock In \emph{Proceedings of the Sixth Workshop on Statistical Machine Translation}, pages 22--64, Edinburgh, Scotland. Association for Computational Linguistics.

\bibitem[{Chowdhery et~al.(2022)Chowdhery, Narang, Devlin, Bosma, Mishra, Roberts, Barham, Chung, Sutton, Gehrmann, Schuh, Shi, Tsvyashchenko, Maynez, Rao, Barnes, Tay, Shazeer, Prabhakaran, Reif, Du, Hutchinson, Pope, Bradbury, Austin, Isard, Gur-Ari, Yin, Duke, Levskaya, Ghemawat, Dev, Michalewski, Garcia, Misra, Robinson, Fedus, Zhou, Ippolito, Luan, Lim, Zoph, Spiridonov, Sepassi, Dohan, Agrawal, Omernick, Dai, Pillai, Pellat, Lewkowycz, Moreira, Child, Polozov, Lee, Zhou, Wang, Saeta, Diaz, Firat, Catasta, Wei, Meier-Hellstern, Eck, Dean, Petrov, and Fiedel}]{chowdhery2022palm}
Aakanksha Chowdhery, Sharan Narang, Jacob Devlin, Maarten Bosma, Gaurav Mishra, Adam Roberts, Paul Barham, Hyung~Won Chung, Charles Sutton, Sebastian Gehrmann, Parker Schuh, Kensen Shi, Sasha Tsvyashchenko, Joshua Maynez, Abhishek Rao, Parker Barnes, Yi~Tay, Noam Shazeer, Vinodkumar Prabhakaran, Emily Reif, Nan Du, Ben Hutchinson, Reiner Pope, James Bradbury, Jacob Austin, Michael Isard, Guy Gur-Ari, Pengcheng Yin, Toju Duke, Anselm Levskaya, Sanjay Ghemawat, Sunipa Dev, Henryk Michalewski, Xavier Garcia, Vedant Misra, Kevin Robinson, Liam Fedus, Denny Zhou, Daphne Ippolito, David Luan, Hyeontaek Lim, Barret Zoph, Alexander Spiridonov, Ryan Sepassi, David Dohan, Shivani Agrawal, Mark Omernick, Andrew~M. Dai, Thanumalayan~Sankaranarayana Pillai, Marie Pellat, Aitor Lewkowycz, Erica Moreira, Rewon Child, Oleksandr Polozov, Katherine Lee, Zongwei Zhou, Xuezhi Wang, Brennan Saeta, Mark Diaz, Orhan Firat, Michele Catasta, Jason Wei, Kathy Meier-Hellstern, Douglas Eck, Jeff Dean, Slav Petrov, and Noah Fiedel. 2022.
\newblock \href {http://arxiv.org/abs/2204.02311} {Palm: Scaling language modeling with pathways}.

\bibitem[{Freitag et~al.(2022)Freitag, Rei, Mathur, Lo, Stewart, Avramidis, Kocmi, Foster, Lavie, and Martins}]{freitag-etal-2022-results}
Markus Freitag, Ricardo Rei, Nitika Mathur, Chi-kiu Lo, Craig Stewart, Eleftherios Avramidis, Tom Kocmi, George Foster, Alon Lavie, and Andr{\'e} F.~T. Martins. 2022.
\newblock \href {https://aclanthology.org/2022.wmt-1.2} {Results of {WMT}22 metrics shared task: Stop using {BLEU} {--} neural metrics are better and more robust}.
\newblock In \emph{Proceedings of the Seventh Conference on Machine Translation (WMT)}, pages 46--68, Abu Dhabi, United Arab Emirates (Hybrid). Association for Computational Linguistics.

\bibitem[{Gao et~al.(2023{\natexlab{a}})Gao, Zhang, He, Wu, and Wang}]{gao-etal-2023-improving}
Pengzhi Gao, Liwen Zhang, Zhongjun He, Hua Wu, and Haifeng Wang. 2023{\natexlab{a}}.
\newblock \href {https://doi.org/10.18653/v1/2023.findings-acl.766} {Improving zero-shot multilingual neural machine translation by leveraging cross-lingual consistency regularization}.
\newblock In \emph{Findings of the Association for Computational Linguistics: ACL 2023}, pages 12103--12119, Toronto, Canada. Association for Computational Linguistics.

\bibitem[{Gao et~al.(2023{\natexlab{b}})Gao, Zhang, He, Wu, and Wang}]{gao-etal-2023-learning-multilingual}
Pengzhi Gao, Liwen Zhang, Zhongjun He, Hua Wu, and Haifeng Wang. 2023{\natexlab{b}}.
\newblock \href {https://doi.org/10.18653/v1/2023.emnlp-industry.25} {Learning multilingual sentence representations with cross-lingual consistency regularization}.
\newblock In \emph{Proceedings of the 2023 Conference on Empirical Methods in Natural Language Processing: Industry Track}, pages 243--262, Singapore. Association for Computational Linguistics.

\bibitem[{Hendy et~al.(2023)Hendy, Abdelrehim, Sharaf, Raunak, Gabr, Matsushita, Kim, Afify, and Awadalla}]{hendy2023good}
Amr Hendy, Mohamed Abdelrehim, Amr Sharaf, Vikas Raunak, Mohamed Gabr, Hitokazu Matsushita, Young~Jin Kim, Mohamed Afify, and Hany~Hassan Awadalla. 2023.
\newblock \href {http://arxiv.org/abs/2302.09210} {How good are gpt models at machine translation? a comprehensive evaluation}.

\bibitem[{Hinton and Roweis(2002)}]{NIPS2002_6150ccc6}
Geoffrey~E Hinton and Sam Roweis. 2002.
\newblock \href {https://proceedings.neurips.cc/paper/2002/file/6150ccc6069bea6b5716254057a194ef-Paper.pdf} {Stochastic neighbor embedding}.
\newblock In \emph{Advances in Neural Information Processing Systems}, volume~15. MIT Press.

\bibitem[{Hu et~al.(2022)Hu, Shen, Wallis, Allen-Zhu, Li, Wang, Wang, and Chen}]{hu2022lora}
Edward~J Hu, Yelong Shen, Phillip Wallis, Zeyuan Allen-Zhu, Yuanzhi Li, Shean Wang, Lu~Wang, and Weizhu Chen. 2022.
\newblock \href {https://openreview.net/forum?id=nZeVKeeFYf9} {Lo{RA}: Low-rank adaptation of large language models}.
\newblock In \emph{International Conference on Learning Representations}.

\bibitem[{Jiao et~al.(2023)Jiao, Wang, tse Huang, Wang, Shi, and Tu}]{jiao2023chatgpt}
Wenxiang Jiao, Wenxuan Wang, Jen tse Huang, Xing Wang, Shuming Shi, and Zhaopeng Tu. 2023.
\newblock \href {http://arxiv.org/abs/2301.08745} {Is chatgpt a good translator? yes with gpt-4 as the engine}.

\bibitem[{Joulin et~al.(2016)Joulin, Grave, Bojanowski, Douze, Jégou, and Mikolov}]{joulin2016fasttextzip}
Armand Joulin, Edouard Grave, Piotr Bojanowski, Matthijs Douze, Hérve Jégou, and Tomas Mikolov. 2016.
\newblock \href {http://arxiv.org/abs/1612.03651} {Fasttext.zip: Compressing text classification models}.

\bibitem[{Joulin et~al.(2017)Joulin, Grave, Bojanowski, and Mikolov}]{joulin-etal-2017-bag}
Armand Joulin, Edouard Grave, Piotr Bojanowski, and Tomas Mikolov. 2017.
\newblock \href {https://aclanthology.org/E17-2068} {Bag of tricks for efficient text classification}.
\newblock In \emph{Proceedings of the 15th Conference of the {E}uropean Chapter of the Association for Computational Linguistics: Volume 2, Short Papers}, pages 427--431, Valencia, Spain. Association for Computational Linguistics.

\bibitem[{Kocmi et~al.(2022)Kocmi, Bawden, Bojar, Dvorkovich, Federmann, Fishel, Gowda, Graham, Grundkiewicz, Haddow, Knowles, Koehn, Monz, Morishita, Nagata, Nakazawa, Nov{\'a}k, Popel, and Popovi{\'c}}]{kocmi-etal-2022-findings}
Tom Kocmi, Rachel Bawden, Ond{\v{r}}ej Bojar, Anton Dvorkovich, Christian Federmann, Mark Fishel, Thamme Gowda, Yvette Graham, Roman Grundkiewicz, Barry Haddow, Rebecca Knowles, Philipp Koehn, Christof Monz, Makoto Morishita, Masaaki Nagata, Toshiaki Nakazawa, Michal Nov{\'a}k, Martin Popel, and Maja Popovi{\'c}. 2022.
\newblock \href {https://aclanthology.org/2022.wmt-1.1} {Findings of the 2022 conference on machine translation ({WMT}22)}.
\newblock In \emph{Proceedings of the Seventh Conference on Machine Translation (WMT)}, pages 1--45, Abu Dhabi, United Arab Emirates (Hybrid). Association for Computational Linguistics.

\bibitem[{Li et~al.(2023)Li, Zhou, Huang, Cheng, and Chen}]{li2023eliciting}
Jiahuan Li, Hao Zhou, Shujian Huang, Shanbo Cheng, and Jiajun Chen. 2023.
\newblock \href {http://arxiv.org/abs/2305.15083} {Eliciting the translation ability of large language models via multilingual finetuning with translation instructions}.

\bibitem[{Lin et~al.(2022)Lin, Mihaylov, Artetxe, Wang, Chen, Simig, Ott, Goyal, Bhosale, Du, Pasunuru, Shleifer, Koura, Chaudhary, O{'}Horo, Wang, Zettlemoyer, Kozareva, Diab, Stoyanov, and Li}]{lin-etal-2022-shot}
Xi~Victoria Lin, Todor Mihaylov, Mikel Artetxe, Tianlu Wang, Shuohui Chen, Daniel Simig, Myle Ott, Naman Goyal, Shruti Bhosale, Jingfei Du, Ramakanth Pasunuru, Sam Shleifer, Punit~Singh Koura, Vishrav Chaudhary, Brian O{'}Horo, Jeff Wang, Luke Zettlemoyer, Zornitsa Kozareva, Mona Diab, Veselin Stoyanov, and Xian Li. 2022.
\newblock \href {https://doi.org/10.18653/v1/2022.emnlp-main.616} {Few-shot learning with multilingual generative language models}.
\newblock In \emph{Proceedings of the 2022 Conference on Empirical Methods in Natural Language Processing}, pages 9019--9052, Abu Dhabi, United Arab Emirates. Association for Computational Linguistics.

\bibitem[{OpenAI(2023{\natexlab{a}})}]{openai2023chatgpt}
OpenAI. 2023{\natexlab{a}}.
\newblock Chatgpt [large language model]. https://chat.openai.com/.

\bibitem[{OpenAI(2023{\natexlab{b}})}]{openai2023gpt4}
OpenAI. 2023{\natexlab{b}}.
\newblock \href {http://arxiv.org/abs/2303.08774} {Gpt-4 technical report}.

\bibitem[{Post(2018)}]{post-2018-call}
Matt Post. 2018.
\newblock \href {https://doi.org/10.18653/v1/W18-6319} {A call for clarity in reporting {BLEU} scores}.
\newblock In \emph{Proceedings of the Third Conference on Machine Translation: Research Papers}, pages 186--191, Brussels, Belgium. Association for Computational Linguistics.

\bibitem[{Rei et~al.(2020)Rei, Stewart, Farinha, and Lavie}]{rei-etal-2020-comet}
Ricardo Rei, Craig Stewart, Ana~C Farinha, and Alon Lavie. 2020.
\newblock \href {https://doi.org/10.18653/v1/2020.emnlp-main.213} {{COMET}: A neural framework for {MT} evaluation}.
\newblock In \emph{Proceedings of the 2020 Conference on Empirical Methods in Natural Language Processing (EMNLP)}, pages 2685--2702, Online. Association for Computational Linguistics.

\bibitem[{Taori et~al.(2023)Taori, Gulrajani, Zhang, Dubois, Li, Guestrin, Liang, and Hashimoto}]{alpaca}
Rohan Taori, Ishaan Gulrajani, Tianyi Zhang, Yann Dubois, Xuechen Li, Carlos Guestrin, Percy Liang, and Tatsunori~B. Hashimoto. 2023.
\newblock Stanford alpaca: An instruction-following llama model.
\newblock \url{https://github.com/tatsu-lab/stanford_alpaca}.

\bibitem[{Team et~al.(2022)Team, Costa-jussà, Cross, Çelebi, Elbayad, Heafield, Heffernan, Kalbassi, Lam, Licht, Maillard, Sun, Wang, Wenzek, Youngblood, Akula, Barrault, Gonzalez, Hansanti, Hoffman, Jarrett, Sadagopan, Rowe, Spruit, Tran, Andrews, Ayan, Bhosale, Edunov, Fan, Gao, Goswami, Guzmán, Koehn, Mourachko, Ropers, Saleem, Schwenk, and Wang}]{nllbteam2022language}
NLLB Team, Marta~R. Costa-jussà, James Cross, Onur Çelebi, Maha Elbayad, Kenneth Heafield, Kevin Heffernan, Elahe Kalbassi, Janice Lam, Daniel Licht, Jean Maillard, Anna Sun, Skyler Wang, Guillaume Wenzek, Al~Youngblood, Bapi Akula, Loic Barrault, Gabriel~Mejia Gonzalez, Prangthip Hansanti, John Hoffman, Semarley Jarrett, Kaushik~Ram Sadagopan, Dirk Rowe, Shannon Spruit, Chau Tran, Pierre Andrews, Necip~Fazil Ayan, Shruti Bhosale, Sergey Edunov, Angela Fan, Cynthia Gao, Vedanuj Goswami, Francisco Guzmán, Philipp Koehn, Alexandre Mourachko, Christophe Ropers, Safiyyah Saleem, Holger Schwenk, and Jeff Wang. 2022.
\newblock \href {http://arxiv.org/abs/2207.04672} {No language left behind: Scaling human-centered machine translation}.

\bibitem[{Team(2024)}]{unbabel2024tower}
Unbabel~Research Team. 2024.
\newblock \href {https://unbabel.com/announcing-tower-an-open-multilingual-llm-for-translation-related-tasks/} {Announcing tower: An open multilingual llm for translation-related tasks}.

\bibitem[{Tiedemann(2012)}]{TIEDEMANN12.463}
Jörg Tiedemann. 2012.
\newblock Parallel data, tools and interfaces in opus.
\newblock In \emph{Proceedings of the Eight International Conference on Language Resources and Evaluation (LREC'12)}, Istanbul, Turkey. European Language Resources Association (ELRA).

\bibitem[{Touvron et~al.(2023)Touvron, Martin, Stone, Albert, Almahairi, Babaei, Bashlykov, Batra, Bhargava, Bhosale, Bikel, Blecher, Ferrer, Chen, Cucurull, Esiobu, Fernandes, Fu, Fu, Fuller, Gao, Goswami, Goyal, Hartshorn, Hosseini, Hou, Inan, Kardas, Kerkez, Khabsa, Kloumann, Korenev, Koura, Lachaux, Lavril, Lee, Liskovich, Lu, Mao, Martinet, Mihaylov, Mishra, Molybog, Nie, Poulton, Reizenstein, Rungta, Saladi, Schelten, Silva, Smith, Subramanian, Tan, Tang, Taylor, Williams, Kuan, Xu, Yan, Zarov, Zhang, Fan, Kambadur, Narang, Rodriguez, Stojnic, Edunov, and Scialom}]{touvron2023llama}
Hugo Touvron, Louis Martin, Kevin Stone, Peter Albert, Amjad Almahairi, Yasmine Babaei, Nikolay Bashlykov, Soumya Batra, Prajjwal Bhargava, Shruti Bhosale, Dan Bikel, Lukas Blecher, Cristian~Canton Ferrer, Moya Chen, Guillem Cucurull, David Esiobu, Jude Fernandes, Jeremy Fu, Wenyin Fu, Brian Fuller, Cynthia Gao, Vedanuj Goswami, Naman Goyal, Anthony Hartshorn, Saghar Hosseini, Rui Hou, Hakan Inan, Marcin Kardas, Viktor Kerkez, Madian Khabsa, Isabel Kloumann, Artem Korenev, Punit~Singh Koura, Marie-Anne Lachaux, Thibaut Lavril, Jenya Lee, Diana Liskovich, Yinghai Lu, Yuning Mao, Xavier Martinet, Todor Mihaylov, Pushkar Mishra, Igor Molybog, Yixin Nie, Andrew Poulton, Jeremy Reizenstein, Rashi Rungta, Kalyan Saladi, Alan Schelten, Ruan Silva, Eric~Michael Smith, Ranjan Subramanian, Xiaoqing~Ellen Tan, Binh Tang, Ross Taylor, Adina Williams, Jian~Xiang Kuan, Puxin Xu, Zheng Yan, Iliyan Zarov, Yuchen Zhang, Angela Fan, Melanie Kambadur, Sharan Narang, Aurelien Rodriguez, Robert Stojnic, Sergey Edunov, and Thomas
  Scialom. 2023.
\newblock \href {http://arxiv.org/abs/2307.09288} {Llama 2: Open foundation and fine-tuned chat models}.

\bibitem[{Vaswani et~al.(2017)Vaswani, Shazeer, Parmar, Uszkoreit, Jones, Gomez, Kaiser, and Polosukhin}]{NIPS2017_3f5ee243}
Ashish Vaswani, Noam Shazeer, Niki Parmar, Jakob Uszkoreit, Llion Jones, Aidan~N Gomez, \L~ukasz Kaiser, and Illia Polosukhin. 2017.
\newblock \href {https://proceedings.neurips.cc/paper_files/paper/2017/file/3f5ee243547dee91fbd053c1c4a845aa-Paper.pdf} {Attention is all you need}.
\newblock In \emph{Advances in Neural Information Processing Systems}, volume~30. Curran Associates, Inc.

\bibitem[{Vilar et~al.(2023)Vilar, Freitag, Cherry, Luo, Ratnakar, and Foster}]{vilar-etal-2023-prompting}
David Vilar, Markus Freitag, Colin Cherry, Jiaming Luo, Viresh Ratnakar, and George Foster. 2023.
\newblock \href {https://doi.org/10.18653/v1/2023.acl-long.859} {Prompting {P}a{LM} for translation: Assessing strategies and performance}.
\newblock In \emph{Proceedings of the 61st Annual Meeting of the Association for Computational Linguistics (Volume 1: Long Papers)}, pages 15406--15427, Toronto, Canada. Association for Computational Linguistics.

\bibitem[{Wu et~al.(2021)Wu, Cheng, Wang, and Li}]{wu-etal-2021-language}
Liwei Wu, Shanbo Cheng, Mingxuan Wang, and Lei Li. 2021.
\newblock \href {https://doi.org/10.18653/v1/2021.findings-acl.264} {Language tags matter for zero-shot neural machine translation}.
\newblock In \emph{Findings of the Association for Computational Linguistics: ACL-IJCNLP 2021}, pages 3001--3007, Online. Association for Computational Linguistics.

\bibitem[{Xu et~al.(2023)Xu, Kim, Sharaf, and Awadalla}]{xu2023paradigm}
Haoran Xu, Young~Jin Kim, Amr Sharaf, and Hany~Hassan Awadalla. 2023.
\newblock \href {http://arxiv.org/abs/2309.11674} {A paradigm shift in machine translation: Boosting translation performance of large language models}.

\bibitem[{Yang et~al.(2023)Yang, Li, Zhang, and Zong}]{yang2023bigtranslate}
Wen Yang, Chong Li, Jiajun Zhang, and Chengqing Zong. 2023.
\newblock \href {http://arxiv.org/abs/2305.18098} {Bigtranslate: Augmenting large language models with multilingual translation capability over 100 languages}.

\bibitem[{Zeng et~al.(2023)Zeng, Meng, Yin, and Zhou}]{zeng2023tim}
Jiali Zeng, Fandong Meng, Yongjing Yin, and Jie Zhou. 2023.
\newblock \href {http://arxiv.org/abs/2307.04408} {Tim: Teaching large language models to translate with comparison}.

\bibitem[{Zhang et~al.(2023{\natexlab{a}})Zhang, Haddow, and Birch}]{zhang2023prompting}
Biao Zhang, Barry Haddow, and Alexandra Birch. 2023{\natexlab{a}}.
\newblock Prompting large language model for machine translation: A case study.
\newblock In \emph{Proceedings of the 40th International Conference on Machine Learning}, ICML'23. JMLR.org.

\bibitem[{Zhang et~al.(2023{\natexlab{b}})Zhang, Fang, Zhang, Ma, Zhou, Huang, Bu, Gui, Chen, Chen, and Feng}]{zhang2023bayling}
Shaolei Zhang, Qingkai Fang, Zhuocheng Zhang, Zhengrui Ma, Yan Zhou, Langlin Huang, Mengyu Bu, Shangtong Gui, Yunji Chen, Xilin Chen, and Yang Feng. 2023{\natexlab{b}}.
\newblock \href {http://arxiv.org/abs/2306.10968} {Bayling: Bridging cross-lingual alignment and instruction following through interactive translation for large language models}.

\bibitem[{Zhu et~al.(2023)Zhu, Liu, Dong, Xu, Huang, Kong, Chen, and Li}]{zhu2023multilingual}
Wenhao Zhu, Hongyi Liu, Qingxiu Dong, Jingjing Xu, Shujian Huang, Lingpeng Kong, Jiajun Chen, and Lei Li. 2023.
\newblock \href {http://arxiv.org/abs/2304.04675} {Multilingual machine translation with large language models: Empirical results and analysis}.

\end{thebibliography}

\appendix

\section{Statistics of All Training Datasets}

\begin{table}[H]
\centering
\begin{tabular}{c | c | c | c} 
\texttt{en} $\leftrightarrow$ & \#sentences & \texttt{en} $\leftrightarrow$ & \#sentences \\
\hline
\hline
\texttt{cs} & $22128$ & \texttt{ru} & $27976$ \\
\texttt{de} & $26398$ & \texttt{zh} & $28788$ \\
\hline
\end{tabular}
\caption{Statistics of the training datasets. Each entry shows the total number of parallel sentence pairs for both directions. Note that \texttt{en} $\rightarrow$ and \texttt{en} $\leftarrow$ directions have an equal number of sentence pairs.}
\label{tab:dataset}
\end{table}

\begin{table}[H]
\centering
\begin{tabular}{c | c | c | c} 
 & \#sentences &  & \#sentences \\
\hline
\hline
\texttt{en}$\leftarrow$\texttt{de} & $23071$ & \texttt{en}$\rightarrow$\texttt{de} & $27175$ \\
\texttt{en}$\leftarrow$\texttt{es} & $1000$ & \texttt{en}$\rightarrow$\texttt{es} & $6708$ \\
\texttt{en}$\leftarrow$\texttt{fr} & $3896$ & \texttt{en}$\rightarrow$\texttt{fr} & $7883$ \\
\texttt{en}$\leftarrow$\texttt{it} & $1000$ & \texttt{en}$\rightarrow$\texttt{it} & $5070$ \\
\texttt{en}$\leftarrow$\texttt{ko} & $1000$ & \texttt{en}$\rightarrow$\texttt{ko} & $2782$ \\
\texttt{en}$\leftarrow$\texttt{nl} & $1000$ & \texttt{en}$\rightarrow$\texttt{nl} & $3908$ \\
\texttt{en}$\leftarrow$\texttt{pt} & $1000$ & \texttt{en}$\rightarrow$\texttt{pt} & $10905$ \\
\texttt{en}$\leftarrow$\texttt{ru} & $15953$ & \texttt{en}$\rightarrow$\texttt{ru} & $22038$ \\
\texttt{en}$\leftarrow$\texttt{zh} & $8518$ & \texttt{en}$\rightarrow$\texttt{zh} & $10521$ \\
\texttt{de}$\leftarrow$\texttt{fr} & $2889$ & \texttt{de}$\rightarrow$\texttt{fr} & $1622$ \\
\hline
\end{tabular}
\caption{Statistics of the translation pairs in TowerBlocks.}
\label{tab:dataset_tower}
\end{table}

\section{Details of Evaluation Results on ALMA Models}

\begin{table*}[h]\scriptsize
\centering
\begin{tabular}{c | c c c c c | c c c c c }
\hline
 & \multicolumn{5}{c|}{T-ENC (full-weight)} & \multicolumn{5}{c}{T-ENC (LoRA)} \\
\hline
 & \texttt{en} & \texttt{cs} & \texttt{de} & \texttt{ru} & \texttt{zh} & \texttt{en} & \texttt{cs} & \texttt{de} & \texttt{ru} & \texttt{zh} \\
\hline
\texttt{en}$\rightarrow$ & - & 27.5/88.7 & 30.2/85.3 & 26.3/86.7 & 37.0/84.9 & - & 29.3/88.7 & 29.3/85.1 & 26.8/86.5 & 35.5/84.3 \\
\texttt{cs}$\rightarrow$ & 42.2/85.7 & - & 23.1/85.3 & 21.1/86.4 & 29.0/82.8 & 44.1/86.0 & - & 24.0/86.0 & 22.6/88.7 & 30.2/83.6 \\
\texttt{de}$\rightarrow$ & 29.9/83.9 & 22.0/90.5 & - & 21.5/87.0 & 31.8/84.8 & 30.1/83.7 & 22.7/90.7 & - & 22.0/87.9 & 31.0/84.3 \\
\texttt{ru}$\rightarrow$ & 37.9/84.7 & 19.4/89.5 & 21.0/83.1 & - & 30.2/83.3 & 40.3/84.7 & 20.0/89.7 & 21.5/83.4 & - & 29.8/83.2 \\
\texttt{zh}$\rightarrow$ & 21.9/78.7 & 11.7/84.2 & 13.9/80.6 & 10.3/76.4 & - & 21.1/78.0 & 10.9/83.0 & 16.1/82.6 & 12.2/81.8 & - \\
\hline
\hline
 & \multicolumn{5}{c|}{T-DEC (full-weight)} & \multicolumn{5}{c}{T-DEC (LoRA)} \\
\hline
 & \texttt{en} & \texttt{cs} & \texttt{de} & \texttt{ru} & \texttt{zh} & \texttt{en} & \texttt{cs} & \texttt{de} & \texttt{ru} & \texttt{zh} \\
\hline
\texttt{en}$\rightarrow$ & - & 28.3/88.7 & 30.7/85.4 & 26.3/86.9 & 37.5/85.2 & - & 29.9/89.3 & 29.8/85.5 & 26.9/86.8 & 36.3/85.0 \\
\texttt{cs}$\rightarrow$ & 41.7/85.4 & - & 24.1/86.3 & 22.0/88.8 & 30.7/84.4 & 44.2/86.1 & - & 24.1/86.3 & 22.3/88.7 & 30.0/83.7 \\
\texttt{de}$\rightarrow$ & 29.8/84.0 & 22.0/90.7 & - & 21.5/88.1 & 31.3/84.8 & 30.3/84.2 & 22.9/90.6 & - & 21.9/87.7 & 31.4/84.6 \\
\texttt{ru}$\rightarrow$ & 37.7/84.7 & 20.2/89.7 & 21.7/83.3 & - & 30.3/83.8 & 40.2/85.0 & 20.3/89.7 & 21.9/83.5 & - & 29.9/83.2 \\
\texttt{zh}$\rightarrow$ & 21.9/78.9 & 14.2/88.0 & 16.1/83.0 & 14.8/87.0 & - & 22.7/79.8 & 13.7/87.4 & 16.3/83.0 & 13.7/84.6 & - \\
\hline
\hline
 & \multicolumn{5}{c|}{S-ENC-T-ENC (full-weight)} & \multicolumn{5}{c}{S-ENC-T-ENC (LoRA)} \\
\hline
 & \texttt{en} & \texttt{cs} & \texttt{de} & \texttt{ru} & \texttt{zh} & \texttt{en} & \texttt{cs} & \texttt{de} & \texttt{ru} & \texttt{zh} \\
\hline
\texttt{en}$\rightarrow$ & - & 27.4/88.7 & 30.1/85.3 & 26.2/86.9 & 36.4/84.7 & - & 29.7/89.0 & 29.6/85.1 & 26.7/86.8 & 35.8/84.4 \\
\texttt{cs}$\rightarrow$ & 41.8/85.5 & - & 6.8/67.1 & 1.5/48.4 & 0.7/60.6 & 44.0/86.0 & - & 24.2/86.1 & 20.9/87.3 & 16.2/75.7 \\
\texttt{de}$\rightarrow$ & 30.2/83.9 & 3.9/62.3 & - & 2.0/49.4 & 1.8/62.0 & 30.1/84.0 & 22.3/90.0 & - & 21.6/87.8 & 21.9/79.2 \\
\texttt{ru}$\rightarrow$ & 38.4/84.8 & 1.5/58.1 & 4.6/63.6 & - & 0.2/59.5 & 39.9/84.9 & 20.1/89.1 & 21.5/83.0 & - & 23.8/79.5 \\
\texttt{zh}$\rightarrow$ & 22.2/78.7 & 1.2/58.9 & 4.6/66.2 & 0.6/48.3 & - & 21.6/78.2 & 11.9/83.4 & 15.6/82.3 & 13.0/83.7 & - \\
\hline
\hline
 & \multicolumn{5}{c|}{S-ENC-T-DEC (full-weight)} & \multicolumn{5}{c}{S-ENC-T-DEC (LoRA)} \\
\hline
 & \texttt{en} & \texttt{cs} & \texttt{de} & \texttt{ru} & \texttt{zh} & \texttt{en} & \texttt{cs} & \texttt{de} & \texttt{ru} & \texttt{zh} \\
\hline
\texttt{en}$\rightarrow$ & - & 27.6/88.9 & 30.6/85.4 & 26.0/86.9 & 37.3/85.1 & - & 29.8/89.2 & 29.4/85.1 & 26.8/86.8 & 36.0/84.8 \\
\texttt{cs}$\rightarrow$ & 42.0/85.7 & - & 23.9/86.2 & 22.3/88.9 & 31.2/84.5 & 44.2/86.0 & - & 24.3/86.3 & 22.4/88.3 & 30.2/83.7 \\
\texttt{de}$\rightarrow$ & 30.0/84.1 & 22.3/90.8 & - & 22.1/88.1 & 32.0/84.7 & 30.4/84.1 & 22.8/90.5 & - & 21.6/87.4 & 31.3/84.5 \\
\texttt{ru}$\rightarrow$ & 38.0/84.8 & 20.1/89.7 & 22.1/83.3 & - & 30.4/83.5 & 40.2/85.0 & 18.3/87.1 & 22.4/83.5 & - & 25.2/79.7 \\
\texttt{zh}$\rightarrow$ & 22.5/79.3 & 13.6/87.8 & 16.2/82.9 & 12.7/81.1 & - & 23.1/79.7 & 13.5/87.1 & 15.7/81.9 & 12.4/82.4 & - \\
\hline
\hline
 & \multicolumn{5}{c|}{GPT-MT (full-weight)} & \multicolumn{5}{c}{GPT-MT (LoRA)} \\
\hline
 & \texttt{en} & \texttt{cs} & \texttt{de} & \texttt{ru} & \texttt{zh} & \texttt{en} & \texttt{cs} & \texttt{de} & \texttt{ru} & \texttt{zh} \\
\hline
\texttt{en}$\rightarrow$ & - & 27.8/88.7 & 30.7/85.4 & 26.5/86.8 & 36.9/85.0 & - & 29.8/89.1 & 29.9/85.3 & 26.8/86.9 & 36.2/84.7 \\
\texttt{cs}$\rightarrow$ & 41.6/85.7 & - & 23.9/86.1 & 21.6/87.9 & 30.7/84.3 & 43.9/86.0 & - & 24.6/86.2 & 22.5/88.8 & 30.2/83.9 \\
\texttt{de}$\rightarrow$ & 30.0/84.1 & 22.0/90.7 & - & 21.8/88.0 & 31.9/84.9 & 30.4/84.2 & 22.0/89.1 & - & 21.6/87.8 & 31.3/84.7 \\
\texttt{ru}$\rightarrow$ & 37.8/84.7 & 20.2/89.7 & 21.7/83.1 & - & 30.2/83.6 & 40.4/85.0 & 19.7/88.5 & 22.5/83.5 & - & 29.5/83.3 \\
\texttt{zh}$\rightarrow$ & 22.4/79.4 & 13.9/87.8 & 16.1/82.8 & 10.6/76.7 & - & 23.0/79.7 & 10.7/81.3 & 15.8/82.8 & 13.9/85.8 & - \\
\end{tabular}
\caption{Performance (SacreBLEU/COMET) on WMT22 and FLORES-200 benchmarks with full-weight and LoRA vanilla finetuning on ALMA-7B-Pretrain.}
\label{tab:en-centric_7B_vanilla}
\end{table*}

\begin{table*}[h]\scriptsize
\centering
\begin{tabular}{c | c c c c c | c c c c c }
\hline
 & \multicolumn{5}{c|}{T-ENC (full-weight)} & \multicolumn{5}{c}{T-ENC (LoRA)} \\
\hline
 & \texttt{en} & \texttt{cs} & \texttt{de} & \texttt{ru} & \texttt{zh} & \texttt{en} & \texttt{cs} & \texttt{de} & \texttt{ru} & \texttt{zh} \\
\hline
\texttt{en}$\rightarrow$ & - & 29.6/89.4 & 31.1/85.5 & 27.6/87.3 & 39.2/85.7 & - & 32.1/89.7 & 31.3/85.3 & 28.7/87.5 & 39.3/85.8 \\
\texttt{cs}$\rightarrow$ & 43.4/86.0 & - & 24.8/86.1 & 22.7/88.2 & 32.3/84.5 & 47.0/86.6 & - & 24.5/86.2 & 22.8/88.5 & 28.3/83.1 \\
\texttt{de}$\rightarrow$ & 30.9/84.3 & 23.3/91.2 & - & 23.2/88.0 & 33.7/85.4 & 31.1/84.5 & 24.2/91.2 & - & 23.2/87.6 & 32.1/84.3 \\
\texttt{ru}$\rightarrow$ & 38.8/85.0 & 21.3/90.1 & 22.4/83.7 & - & 31.4/84.0 & 41.4/85.4 & 21.4/90.3 & 23.2/83.9 & - & 30.6/83.5 \\
\texttt{zh}$\rightarrow$ & 23.8/80.1 & 15.0/88.3 & 16.0/82.3 & 14.6/83.3 & - & 24.6/80.2 & 13.5/86.2 & 16.5/82.6 & 7.9/73.3 & - \\
\hline
\hline
 & \multicolumn{5}{c|}{T-DEC (full-weight)} & \multicolumn{5}{c}{T-DEC (LoRA)} \\
\hline
 & \texttt{en} & \texttt{cs} & \texttt{de} & \texttt{ru} & \texttt{zh} & \texttt{en} & \texttt{cs} & \texttt{de} & \texttt{ru} & \texttt{zh} \\
\hline
\texttt{en}$\rightarrow$ & - & 29.6/89.4 & 31.6/85.5 & 27.2/87.4 & 38.4/85.7 & - & 32.0/89.8 & 31.3/85.5 & 29.1/87.5 & 39.5/85.7 \\
\texttt{cs}$\rightarrow$ & 43.4/86.1 & - & 25.2/86.7 & 11.6/66.2 & 12.7/69.7 & 46.7/86.7 & - & 25.7/86.5 & 20.6/83.6 & 32.9/84.9 \\
\texttt{de}$\rightarrow$ & 31.5/84.6 & 23.0/91.3 & - & 8.9/61.1 & 33.6/85.4 & 31.2/84.6 & 24.1/91.3 & - & 21.3/83.9 & 33.9/85.4 \\
\texttt{ru}$\rightarrow$ & 38.2/85.2 & 21.2/90.2 & 23.1/83.9 & - & 28.8/81.6 & 40.9/85.4 & 21.7/90.3 & 24.0/83.7 & - & 32.5/84.3 \\
\texttt{zh}$\rightarrow$ & 23.6/80.1 & 15.0/88.7 & 17.1/83.5 & 15.8/87.4 & - & 24.1/80.0 & 15.9/89.0 & 7.0/67.3 & 12.1/76.4 & - \\
\hline
\hline
 & \multicolumn{5}{c|}{S-ENC-T-ENC (full-weight)} & \multicolumn{5}{c}{S-ENC-T-ENC (LoRA)} \\
\hline
 & \texttt{en} & \texttt{cs} & \texttt{de} & \texttt{ru} & \texttt{zh} & \texttt{en} & \texttt{cs} & \texttt{de} & \texttt{ru} & \texttt{zh} \\
\hline
\texttt{en}$\rightarrow$ & - & 30.1/89.5 & 31.2/85.7 & 27.5/87.4 & 39.5/85.7 & - & 32.1/89.7 & 31.1/85.4 & 28.8/87.5 & 39.5/85.9 \\
\texttt{cs}$\rightarrow$ & 43.2/85.9 & - & 8.6/68.2 & 2.0/49.0 & 5.7/65.0 & 47.0/86.5 & - & 6.9/66.8 & 4.4/53.8 & 24.9/79.6 \\
\texttt{de}$\rightarrow$ & 30.9/84.4 & 14.0/78.9 & - & 7.5/60.9 & 23.4/78.7 & 31.4/84.6 & 15.4/79.6 & - & 4.2/53.7 & 32.5/84.6 \\
\texttt{ru}$\rightarrow$ & 38.5/85.0 & 16.0/83.1 & 17.5/78.1 & - & 27.0/80.6 & 41.4/85.4 & 6.4/67.0 & 2.6/61.3 & - & 26.4/80.0 \\
\texttt{zh}$\rightarrow$ & 23.6/80.3 & 2.4/61.8 & 2.7/62.9 & 0.6/48.3 & - & 24.6/80.1 & 1.7/60.2 & 2.3/62.6 & 0.9/49.3 & - \\
\hline
\hline
 & \multicolumn{5}{c|}{S-ENC-T-DEC (full-weight)} & \multicolumn{5}{c}{S-ENC-T-DEC (LoRA)} \\
\hline
 & \texttt{en} & \texttt{cs} & \texttt{de} & \texttt{ru} & \texttt{zh} & \texttt{en} & \texttt{cs} & \texttt{de} & \texttt{ru} & \texttt{zh} \\
\hline
\texttt{en}$\rightarrow$ & - & 29.4/89.2 & 31.2/85.6 & 27.3/87.3 & 38.7/85.6 & - & 32.5/90.0 & 31.6/85.6 & 29.2/87.6 & 39.7/85.8 \\
\texttt{cs}$\rightarrow$ & 43.2/86.1 & - & 24.1/85.2 & 23.3/89.2 & 4.9/63.8 & 46.4/86.6 & - & 6.0/66.0 & 0.9/47.6 & 1.2/61.5 \\
\texttt{de}$\rightarrow$ & 31.1/84.4 & 23.1/91.3 & - & 23.0/88.5 & 33.5/85.4 & 31.3/84.6 & 4.6/64.1 & - & 1.0/48.0 & 31.6/82.6 \\
\texttt{ru}$\rightarrow$ & 38.3/85.1 & 21.1/90.1 & 22.6/83.4 & - & 24.7/77.5 & 40.8/85.4 & 3.5/61.6 & 2.1/60.5 & - & 7.1/64.5 \\
\texttt{zh}$\rightarrow$ & 23.6/80.1 & 15.1/88.0 & 12.9/77.7 & 15.8/87.6 & - & 24.6/80.2 & 1.3/59.5 & 1.6/61.8 & 0.7/48.5 & - \\
\hline
\hline
 & \multicolumn{5}{c|}{GPT-MT (full-weight)} & \multicolumn{5}{c}{GPT-MT (LoRA)} \\
\hline
 & \texttt{en} & \texttt{cs} & \texttt{de} & \texttt{ru} & \texttt{zh} & \texttt{en} & \texttt{cs} & \texttt{de} & \texttt{ru} & \texttt{zh} \\
\hline
\texttt{en}$\rightarrow$ & - & 29.6/89.2 & 31.5/85.4 & 27.7/87.3 & 39.3/85.5 & - & 32.3/89.7 & 31.4/85.7 & 28.9/87.5 & 39.6/85.6 \\
\texttt{cs}$\rightarrow$ & 42.7/86.0 & - & 22.7/84.3 & 1.1/47.4 & 1.8/61.7 & 46.3/86.6 & - & 18.4/78.2 & 2.0/49.2 & 23.1/77.3 \\
\texttt{de}$\rightarrow$ & 30.9/84.6 & 23.4/91.2 & - & 3.2/50.5 & 33.9/85.6 & 31.6/84.7 & 19.3/83.3 & - & 1.9/48.7 & 32.3/84.0 \\
\texttt{ru}$\rightarrow$ & 38.6/85.0 & 21.2/90.1 & 22.6/83.3 & - & 22.7/76.7 & 40.7/85.3 & 14.9/79.4 & 20.0/79.6 & - & 20.9/74.8 \\
\texttt{zh}$\rightarrow$ & 23.5/80.2 & 15.4/88.3 & 13.1/78.2 & 1.4/50.0 & - & 24.4/80.0 & 9.4/75.6 & 5.0/65.7 & 0.7/48.6 & - \\
\end{tabular}
\caption{Performance (SacreBLEU/COMET) on WMT22 and FLORES-200 benchmarks with full-weight and LoRA vanilla finetuning on ALMA-13B-Pretrain.}
\label{tab:en-centric_13B_vanilla}
\end{table*}

\begin{table*}[h]\scriptsize
\centering
\begin{tabular}{c | c c c c c | c c c c c }
\hline
 & \multicolumn{5}{c|}{T-ENC (full-weight)} & \multicolumn{5}{c}{T-ENC (LoRA)} \\
\hline
 & \texttt{en} & \texttt{cs} & \texttt{de} & \texttt{ru} & \texttt{zh} & \texttt{en} & \texttt{cs} & \texttt{de} & \texttt{ru} & \texttt{zh} \\
\hline
\texttt{en}$\rightarrow$ & - & 27.8/88.7 & 30.6/85.4 & 26.2/86.9 & 37.3/85.2 & - & 29.5/88.8 & 29.3/85.2 & 26.6/86.6 & 35.6/84.5 \\
\texttt{cs}$\rightarrow$ & 41.9/85.5 & - & 23.7/86.0 & 22.1/88.6 & 30.7/84.0 & 44.3/86.1 & - & 24.4/86.3 & 22.5/88.8 & 29.5/83.5 \\
\texttt{de}$\rightarrow$ & 29.5/84.1 & 21.9/90.5 & - & 22.2/88.0 & 32.0/84.8 & 29.9/83.7 & 22.1/90.8 & - & 21.9/88.0 & 30.4/84.2 \\
\texttt{ru}$\rightarrow$ & 38.3/84.6 & 19.7/89.7 & 21.5/83.2 & - & 30.5/83.5 & 40.3/84.7 & 19.9/89.6 & 22.4/83.5 & - & 28.3/83.0 \\
\texttt{zh}$\rightarrow$ & 22.1/79.1 & 13.6/87.8 & 16.0/82.8 & 14.3/86.5 & - & 21.5/77.8 & 13.1/87.0 & 15.4/82.5 & 14.2/86.1 & - \\
\hline
\hline
 & \multicolumn{5}{c|}{T-DEC (full-weight)} & \multicolumn{5}{c}{T-DEC (LoRA)} \\
\hline
 & \texttt{en} & \texttt{cs} & \texttt{de} & \texttt{ru} & \texttt{zh} & \texttt{en} & \texttt{cs} & \texttt{de} & \texttt{ru} & \texttt{zh} \\
\hline
\texttt{en}$\rightarrow$ & - & 28.1/88.9 & 30.5/85.5 & 26.0/86.8 & 37.2/85.1 & - & 29.6/89.3 & 30.0/85.7 & 27.0/86.9 & 36.2/85.0 \\
\texttt{cs}$\rightarrow$ & 41.5/85.4 & - & 24.1/86.1 & 22.0/88.8 & 30.4/84.5 & 43.9/86.1 & - & 24.2/86.2 & 22.8/89.0 & 29.4/83.7 \\
\texttt{de}$\rightarrow$ & 29.7/84.1 & 22.0/90.9 & - & 21.7/88.1 & 31.4/84.7 & 30.4/84.2 & 22.3/90.8 & - & 21.2/87.9 & 30.6/84.5 \\
\texttt{ru}$\rightarrow$ & 38.0/84.7 & 19.9/89.8 & 21.8/83.4 & - & 29.9/83.6 & 40.1/85.0 & 19.9/89.7 & 22.3/83.7 & - & 28.5/82.7 \\
\texttt{zh}$\rightarrow$ & 21.9/79.0 & 13.7/87.8 & 16.3/83.0 & 14.8/86.8 & - & 22.5/79.6 & 13.9/88.0 & 15.6/82.7 & 14.4/86.7 & - \\
\hline
\hline
 & \multicolumn{5}{c|}{S-ENC-T-ENC (full-weight)} & \multicolumn{5}{c}{S-ENC-T-ENC (LoRA)} \\
\hline
 & \texttt{en} & \texttt{cs} & \texttt{de} & \texttt{ru} & \texttt{zh} & \texttt{en} & \texttt{cs} & \texttt{de} & \texttt{ru} & \texttt{zh} \\
\hline
\texttt{en}$\rightarrow$ & - & 27.4/88.7 & 30.1/85.4 & 26.1/86.9 & 36.6/84.9 & - & 29.0/89.0 & 29.4/85.4 & 26.8/86.9 & 35.4/84.6 \\
\texttt{cs}$\rightarrow$ & 41.8/85.5 & - & 23.8/86.0 & 21.5/88.1 & 29.8/83.8 & 43.5/85.9 & - & 23.7/85.8 & 22.0/88.8 & 28.1/83.1 \\
\texttt{de}$\rightarrow$ & 29.5/83.8 & 20.6/90.5 & - & 21.3/87.6 & 30.8/84.4 & 29.5/83.9 & 22.2/90.7 & - & 21.1/87.8 & 29.8/84.0 \\
\texttt{ru}$\rightarrow$ & 38.1/84.7 & 18.9/89.2 & 21.9/83.1 & - & 28.1/82.3 & 40.1/84.8 & 20.0/89.6 & 22.3/83.2 & - & 27.9/82.5 \\
\texttt{zh}$\rightarrow$ & 21.9/78.7 & 12.7/87.4 & 15.8/82.5 & 14.1/85.2 & - & 21.0/77.8 & 12.5/86.3 & 15.4/82.2 & 13.5/86.0 & - \\
\hline
\hline
 & \multicolumn{5}{c|}{S-ENC-T-DEC (full-weight)} & \multicolumn{5}{c}{S-ENC-T-DEC (LoRA)} \\
\hline
 & \texttt{en} & \texttt{cs} & \texttt{de} & \texttt{ru} & \texttt{zh} & \texttt{en} & \texttt{cs} & \texttt{de} & \texttt{ru} & \texttt{zh} \\
\hline
\texttt{en}$\rightarrow$ & - & 27.5/88.8 & 30.6/85.5 & 26.3/87.0 & 37.1/85.1 & - & 29.6/89.2 & 29.4/85.2 & 26.9/86.9 & 36.1/84.8 \\
\texttt{cs}$\rightarrow$ & 41.3/85.6 & - & 24.2/86.2 & 22.1/88.9 & 30.5/84.3 & 44.1/86.1 & - & 24.5/86.4 & 22.7/88.9 & 29.8/83.6 \\
\texttt{de}$\rightarrow$ & 30.0/84.0 & 22.1/90.9 & - & 20.1/83.8 & 31.4/84.4 & 30.3/84.1 & 22.8/90.8 & - & 21.7/87.8 & 31.2/84.4 \\
\texttt{ru}$\rightarrow$ & 38.1/84.8 & 20.1/89.7 & 22.0/83.5 & - & 30.1/83.5 & 40.2/85.0 & 20.5/89.8 & 22.7/83.5 & - & 29.6/83.3 \\
\texttt{zh}$\rightarrow$ & 22.2/79.2 & 14.0/87.8 & 16.6/82.9 & 14.8/86.7 & - & 23.0/79.8 & 13.4/88.0 & 15.9/83.0 & 14.1/86.6 & - \\
\hline
\hline
 & \multicolumn{5}{c|}{GPT-MT (full-weight)} & \multicolumn{5}{c}{GPT-MT (LoRA)} \\
\hline
 & \texttt{en} & \texttt{cs} & \texttt{de} & \texttt{ru} & \texttt{zh} & \texttt{en} & \texttt{cs} & \texttt{de} & \texttt{ru} & \texttt{zh} \\
\hline
\texttt{en}$\rightarrow$ & - & 27.6/88.8 & 30.6/85.4 & 26.4/86.9 & 37.6/85.3 & - & 29.4/89.1 & 29.9/85.4 & 27.0/87.0 & 36.2/84.7 \\
\texttt{cs}$\rightarrow$ & 41.3/85.6 & - & 24.2/86.2 & 22.0/88.9 & 30.6/84.4 & 43.9/86.0 & - & 24.6/86.3 & 22.7/88.8 & 30.1/83.8 \\
\texttt{de}$\rightarrow$ & 29.6/84.2 & 22.3/90.9 & - & 21.4/88.1 & 31.9/84.7 & 30.2/84.2 & 22.5/90.5 & - & 21.9/87.9 & 31.0/84.5 \\
\texttt{ru}$\rightarrow$ & 37.9/84.6 & 19.7/89.6 & 21.8/83.3 & - & 30.0/83.5 & 40.1/84.9 & 20.1/89.2 & 22.5/83.5 & - & 29.9/83.5 \\
\texttt{zh}$\rightarrow$ & 21.4/78.9 & 13.7/87.8 & 16.4/82.9 & 14.5/86.2 & - & 22.8/79.6 & 13.5/87.8 & 16.2/83.1 & 14.4/86.7 & - \\
\end{tabular}
\caption{Performance (SacreBLEU/COMET) on WMT22 and FLORES-200 benchmarks with full-weight and LoRA cross-lingual consistency finetuning on ALMA-7B-Pretrain.}
\label{tab:en-centric_7B_xconst}
\end{table*}

\begin{table*}[h]\scriptsize
\centering
\begin{tabular}{c | c c c c c | c c c c c }
\hline
 & \multicolumn{5}{c|}{T-ENC (full-weight)} & \multicolumn{5}{c}{T-ENC (LoRA)} \\
\hline
 & \texttt{en} & \texttt{cs} & \texttt{de} & \texttt{ru} & \texttt{zh} & \texttt{en} & \texttt{cs} & \texttt{de} & \texttt{ru} & \texttt{zh} \\
\hline
\texttt{en}$\rightarrow$ & - & 29.6/89.5 & 31.1/85.6 & 27.2/87.4 & 39.6/85.8 & - & 31.8/89.9 & 31.3/85.7 & 28.9/87.7 & 39.5/85.8 \\
\texttt{cs}$\rightarrow$ & 44.0/86.1 & - & 25.2/86.5 & 23.3/89.1 & 32.6/84.7 & 46.7/86.6 & - & 26.3/86.5 & 24.0/89.3 & 31.8/84.6 \\
\texttt{de}$\rightarrow$ & 31.1/84.5 & 22.7/91.3 & - & 23.2/88.4 & 33.7/85.4 & 31.2/84.5 & 23.1/91.2 & - & 23.8/88.5 & 33.1/85.4 \\
\texttt{ru}$\rightarrow$ & 39.0/85.1 & 20.8/90.1 & 23.5/83.9 & - & 31.8/84.0 & 40.9/85.3 & 21.2/89.9 & 24.0/84.0 & - & 31.1/83.9 \\
\texttt{zh}$\rightarrow$ & 23.8/80.2 & 15.1/88.5 & 16.8/83.2 & 15.5/86.4 & - & 24.3/80.2 & 14.1/87.6 & 17.1/82.7 & 14.6/84.4 & - \\
\hline
\hline
 & \multicolumn{5}{c|}{T-DEC (full-weight)} & \multicolumn{5}{c}{T-DEC (LoRA)} \\
\hline
 & \texttt{en} & \texttt{cs} & \texttt{de} & \texttt{ru} & \texttt{zh} & \texttt{en} & \texttt{cs} & \texttt{de} & \texttt{ru} & \texttt{zh} \\
\hline
\texttt{en}$\rightarrow$ & - & 29.1/89.3 & 30.9/85.6 & 27.2/87.4 & 39.0/85.9 & - & 31.7/90.0 & 31.5/85.8 & 29.0/87.5 & 39.5/85.9 \\
\texttt{cs}$\rightarrow$ & 43.0/85.9 & - & 24.4/86.5 & 23.0/89.1 & 32.3/84.9 & 46.3/86.6 & - & 26.3/86.8 & 23.5/89.0 & 32.7/85.0 \\
\texttt{de}$\rightarrow$ & 31.1/84.5 & 22.7/91.3 & - & 23.4/88.5 & 33.3/85.4 & 31.4/84.6 & 24.2/91.4 & - & 23.5/88.4 & 33.9/85.5 \\
\texttt{ru}$\rightarrow$ & 38.5/85.0 & 20.7/90.2 & 24.1/84.1 & - & 32.1/84.3 & 40.8/85.4 & 21.9/90.2 & 24.3/84.3 & - & 31.9/83.9 \\
\texttt{zh}$\rightarrow$ & 23.4/80.0 & 15.2/88.6 & 16.8/83.5 & 16.2/87.6 & - & 24.2/80.2 & 15.5/88.9 & 17.7/83.6 & 15.9/87.1 & - \\
\hline
\hline
 & \multicolumn{5}{c|}{S-ENC-T-ENC (full-weight)} & \multicolumn{5}{c}{S-ENC-T-ENC (LoRA)} \\
\hline
 & \texttt{en} & \texttt{cs} & \texttt{de} & \texttt{ru} & \texttt{zh} & \texttt{en} & \texttt{cs} & \texttt{de} & \texttt{ru} & \texttt{zh} \\
\hline
\texttt{en}$\rightarrow$ & - & 29.5/89.4 & 31.2/85.7 & 27.3/87.4 & 39.3/86.0 & - & 31.8/90.0 & 31.0/85.6 & 28.6/87.7 & 39.5/85.8 \\
\texttt{cs}$\rightarrow$ & 43.7/86.0 & - & 25.2/86.2 & 22.9/89.0 & 32.0/84.9 & 46.5/86.5 & - & 26.1/86.7 & 23.8/89.3 & 32.2/84.8 \\
\texttt{de}$\rightarrow$ & 31.4/84.5 & 22.9/91.2 & - & 23.2/88.3 & 33.0/85.2 & 31.4/84.6 & 24.0/91.4 & - & 23.9/88.5 & 34.2/85.5 \\
\texttt{ru}$\rightarrow$ & 38.9/85.1 & 20.5/90.0 & 23.3/83.4 & - & 32.0/84.2 & 41.1/85.3 & 21.8/90.0 & 24.0/84.0 & - & 32.3/84.1 \\
\texttt{zh}$\rightarrow$ & 23.8/80.2 & 15.3/88.5 & 16.3/82.9 & 15.8/87.1 & - & 24.5/80.0 & 13.5/85.9 & 16.8/82.9 & 15.3/85.4 & - \\
\hline
\hline
 & \multicolumn{5}{c|}{S-ENC-T-DEC (full-weight)} & \multicolumn{5}{c}{S-ENC-T-DEC (LoRA)} \\
\hline
 & \texttt{en} & \texttt{cs} & \texttt{de} & \texttt{ru} & \texttt{zh} & \texttt{en} & \texttt{cs} & \texttt{de} & \texttt{ru} & \texttt{zh} \\
\hline
\texttt{en}$\rightarrow$ & - & 28.9/89.3 & 31.2/85.7 & 27.3/87.4 & 39.1/85.9 & - & 31.3/90.0 & 31.7/86.0 & 29.1/87.8 & 39.4/85.9 \\
\texttt{cs}$\rightarrow$ & 42.6/85.9 & - & 26.1/86.7 & 23.2/89.3 & 32.5/85.0 & 45.6/86.6 & - & 26.3/86.8 & 23.3/89.2 & 31.9/84.8 \\
\texttt{de}$\rightarrow$ & 30.5/84.4 & 22.2/91.2 & - & 23.3/88.6 & 33.5/85.4 & 31.2/84.6 & 23.8/91.4 & - & 23.5/88.6 & 33.5/85.4 \\
\texttt{ru}$\rightarrow$ & 37.9/84.9 & 20.9/90.3 & 24.5/84.2 & - & 31.9/84.4 & 40.7/85.2 & 21.6/90.0 & 24.4/84.5 & - & 31.5/84.1 \\
\texttt{zh}$\rightarrow$ & 23.6/80.1 & 15.3/88.7 & 17.6/83.9 & 16.5/87.8 & - & 23.9/80.0 & 15.4/88.6 & 16.4/81.3 & 15.8/86.1 & - \\
\hline
\hline
 & \multicolumn{5}{c|}{GPT-MT (full-weight)} & \multicolumn{5}{c}{GPT-MT (LoRA)} \\
\hline
 & \texttt{en} & \texttt{cs} & \texttt{de} & \texttt{ru} & \texttt{zh} & \texttt{en} & \texttt{cs} & \texttt{de} & \texttt{ru} & \texttt{zh} \\
\hline
\texttt{en}$\rightarrow$ & - & 29.2/89.2 & 31.4/85.7 & 27.6/87.4 & 39.2/85.9 & - & 31.8/89.8 & 31.5/85.7 & 28.9/87.5 & 39.6/85.6 \\
\texttt{cs}$\rightarrow$ & 42.6/85.8 & - & 25.4/86.6 & 23.2/89.1 & 32.1/84.9 & 45.8/86.6 & - & 26.1/86.5 & 23.4/88.8 & 31.9/84.8 \\
\texttt{de}$\rightarrow$ & 30.8/84.5 & 22.1/91.1 & - & 22.9/88.4 & 33.6/85.6 & 31.4/84.7 & 23.8/91.4 & - & 23.4/88.2 & 33.2/85.4 \\
\texttt{ru}$\rightarrow$ & 38.4/85.0 & 20.3/90.1 & 24.2/84.0 & - & 31.5/84.3 & 40.5/85.3 & 21.6/89.8 & 24.1/84.4 & - & 31.5/84.0 \\
\texttt{zh}$\rightarrow$ & 23.6/80.0 & 15.4/88.7 & 17.5/83.6 & 16.3/87.6 & - & 23.9/80.1 & 15.0/88.2 & 16.9/82.6 & 15.6/86.2 & - \\
\end{tabular}
\caption{Performance (SacreBLEU/COMET) on WMT22 and FLORES-200 benchmarks with full-weight and LoRA cross-lingual consistency finetuning on ALMA-13B-Pretrain.}
\label{tab:en-centric_13B_xconst}
\end{table*}

\section{Prompt Strategy for GPTs}

\begin{table}[H]\small
\centering
\begin{tabular}{l} 
Prompt \\
\hline
\hline
Translate this from [src] into [tgt]. Do not provide any \\ 
explanations or text apart from the translation.\\
\textbackslash n[src]: <SRC>\textbackslash n[tgt]:
\end{tabular}
\caption{Prompt strategy adopted when using GPTs, where [src] and [tgt] denote the source and target languages, and <SRC> denotes the source sentence.}
\label{tab:gpt_prompt}
\end{table}

\section{More Evaluation Results on Tower Models}

\begin{table}[H]\small
\centering
\begin{tabular}{l | c | c | c} 
\multicolumn{1}{c|}{Method} & Data & English-centric & Others \\
\hline
\hline
TowerInstruct & \circled{1} & 37.0 / 88.6 & 24.6 / 86.1 \\
\hline
T-ENC & \circled{2} & 37.4 / 88.6 & 24.2 / 85.5 \\
\ \ + XConST & \circled{2} & 37.4 / 88.6 & 25.1 / 86.0 \\
\hdashline
T-DEC & \circled{2} & 37.6 / 88.6 & 24.7 / 86.1 \\
\ \ + XConST & \circled{2} & 37.5 / 88.6 & 25.3 / 86.3 \\
\hdashline
S-ENC-T-ENC & \circled{2} & 37.5 / 88.6 & 24.5 / 85.9 \\
\ \ + XConST & \circled{2} & 37.5 / 88.7 & 25.3 / 86.3 \\
\hdashline
S-ENC-T-DEC & \circled{2} & 37.7 / 88.7 & 24.6 / 86.0 \\
\ \ + XConST & \circled{2} & 37.7 / 88.7 & 25.6 / 86.5 \\
\hdashline
GPT-MT & \circled{2} & 37.9 / 88.7 & 24.8 / 86.2 \\
\ \ + XConST & \circled{2} & 37.8 / 88.8 & 25.6 / 86.5 \\
\end{tabular}
\caption{Performance (SacreBLEU / COMET) on FLORES-200 benchmark with LoRA finetuning on TowerBase-7B. \circled{1} denotes the TowerBlocks dataset. \circled{2} denotes all translation pairs extracted from TowerBlock.}
\label{tab:tower_results_beyond}
\end{table}

\end{document}